\documentclass[a4paper,fleqn]{cas-sc}

\usepackage[authoryear,longnamesfirst]{natbib}
\usepackage{pifont}
\usepackage{lineno}
\usepackage{caption}
\usepackage{tabularx} 
\usepackage{amsmath,amssymb}
\usepackage{algorithm,algpseudocode}
\usepackage{siunitx}             
\usepackage{float}               
\usepackage{pgfplots}
\pgfplotsset{compat=1.18}
\usepackage{empheq}
\usepackage{xcolor}

\usepackage[most]{tcolorbox}
\usepackage{needspace}
\usepackage{listings}
\usepackage{enumitem}

\usepackage{booktabs}
\usepackage{array}
\usepackage{multirow}
\usepackage{xltabular}
\usepackage{ltablex}
\keepXColumns

\newcolumntype{M}[1]{>{\raggedright\arraybackslash}m{#1}}
\newcolumntype{Y}{>{\raggedright\arraybackslash}X}

\lstdefinestyle{jsonstyle}{
    basicstyle=\ttfamily\small,
    breaklines=true,
    columns=fullflexible,
    keepspaces=true,
    frame=none,
    showstringspaces=false
}

\newtcolorbox{promptbox}[2][]{
  enhanced jigsaw,
  breakable,
  sharp corners,
  colback=white,
  colframe=black!35,
  colbacktitle=black!8,
  coltitle=black,
  fonttitle=\bfseries,
  boxrule=0.5pt,
  left=6pt,
  right=6pt,
  top=6pt,
  bottom=6pt,
  title=#2,
  before skip=8pt,
  after skip=10pt,
  pad at break*=1mm,
  segmentation hidden,
  #1
}

\newcommand{\cmark}{\textcolor{green!55!black}{\ding{51}}}
\newcommand{\xmark}{\textcolor{red!70!black}{\ding{55}}}
\newcommand{\dashmark}{\textcolor{black!65}{--}}

\newcommand{\tabciteDriveMLM}{Cui et al., 2025}
\newcommand{\tabciteWaymoQA}{Yu et al., 2025}

\def\tsc#1{\csdef{#1}{\textsc{\lowercase{#1}}\xspace}}
\tsc{WGM}
\tsc{QE}
\tsc{EP}
\tsc{PMS}
\tsc{BEC}
\tsc{DE}

\begin{document}
\let\WriteBookmarks\relax
\def\floatpagepagefraction{1}
\def\textpagefraction{.001}

\shorttitle{V2X-QA: A Benchmark for MLLMs in Autonomous Driving Across Ego, Infrastructure, and Cooperative Views}

\shortauthors{Junwei You et~al.}

\title [mode = title]{V2X-QA: A Comprehensive Reasoning Dataset and Benchmark for Multimodal Large Language Models in Autonomous Driving Across Ego, Infrastructure, and Cooperative Views}                      

\author[1]{Junwei You}[orcid=0009-0002-6447-8276]

\author[2]{Pei Li}
\ead{pei.li@uwyo.edu}
\cormark[1]

\author[3]{Zhuoyu Jiang}

\author[1]{Weizhe Tang}

\author[1]{Zilin Huang}

\author[1]{Rui Gan}

\author[1]{Jiaxi Liu}

\author[4]{Yan Zhao}

\author[1]{Sikai Chen}

\author[1]{Bin Ran}


\affiliation[1]{organization={Department of Civil and Environmental Engineering, University of Wisconsin–Madison},
    city={Madison},
    state={WI},
    postcode={53706}, 
    country={USA}}

\affiliation[2]{organization={Department of Civil and Architectural Engineering and Construction Management, University of Wyoming},
    city={Laramie},
    state={WY},
    postcode={82071}, 
    country={USA}}

\affiliation[3]{organization={Kuaishou Technology},
    city={Beijing},
    postcode={100085},
    country={China}}

\affiliation[4]{organization={School of Transportation, Southeast University},
    city={Nanjing},
    state={Jiangsu},
    postcode={211189}, 
    country={China}}


\cortext[cor1]{Corresponding author}

\begin{abstract}
Multimodal large language models (MLLMs) have shown strong potential for autonomous driving, yet existing benchmarks remain largely ego-centric and therefore cannot systematically assess model performance in infrastructure-centric and cooperative driving conditions. In this work, we introduce V2X-QA, a real-world dataset and benchmark for evaluating MLLMs across vehicle-side, infrastructure-side, and cooperative viewpoints. V2X-QA is built around a view-decoupled evaluation protocol that enables controlled comparison under vehicle-only, infrastructure-only, and cooperative driving conditions within a unified multiple-choice question answering (MCQA) framework. The benchmark is organized into a twelve-task taxonomy spanning perception, prediction, and reasoning and planning, and is constructed through expert-verified MCQA annotation to enable fine-grained diagnosis of viewpoint-dependent capabilities. Benchmark results across ten representative state-of-the-art proprietary and open-source models show that viewpoint accessibility substantially affects performance, and infrastructure-side reasoning supports meaningful macroscopic traffic understanding. Results also indicate that cooperative reasoning remains challenging since it requires cross-view alignment and evidence integration rather than simply additional visual input. To address these challenges, we introduce V2X-MoE, a benchmark-aligned baseline with explicit view routing and viewpoint-specific LoRA experts. The strong performance of V2X-MoE further suggests that explicit viewpoint specialization is a promising direction for multi-view reasoning in autonomous driving. Overall, V2X-QA provides a foundation for studying multi-perspective reasoning, reliability, and cooperative physical intelligence in connected autonomous driving. The dataset and V2X-MoE resources are publicly available at: {\color {blue} https://github.com/junwei0001/V2X-QA}.
\end{abstract}


\maketitle

\section{Introduction}
\label{sec:intro}

\begin{figure}[pos=htbp]
    \centering
    \includegraphics[width=\linewidth]{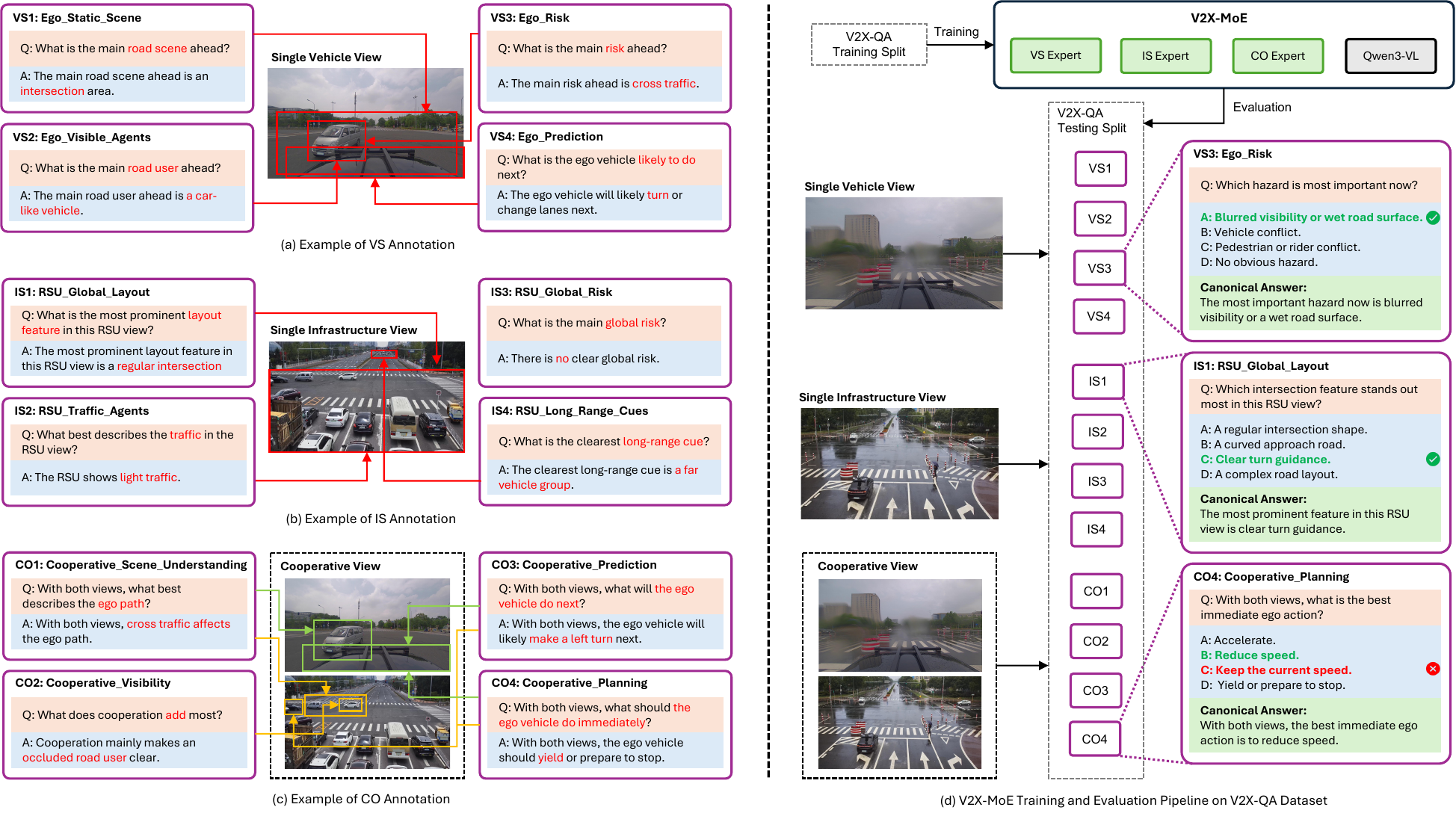}
    \caption{Overview of V2X-QA. Left: representative examples of the twelve viewpoint-aligned tasks under vehicle-side, infrastructure-side, and cooperative settings. Right: MCQA-based training and evaluation pipeline of V2X-MoE on the V2X-QA training and testing splits.}
    \label{fig:v2xqa_overview}
\end{figure}


Multimodal large language models (MLLMs) are increasingly considered as a foundation for semantic understanding and high-level reasoning in autonomous driving. Unlike conventional autonomy stacks that primarily output structured representations for perception and motion, such as detected objects and planned trajectories, these models can also answer questions about traffic scenes, generate grounded natural-language descriptions, and provide interpretable explanations that support debugging, human interaction, and decision-relevant reasoning (\cite{sima2024drivelm, xu2024drivegpt4, gao2024vista, tian2024drivevlm, hwang2024emma}). Such reasoning can complement structured outputs by providing contextual semantics and safety-relevant cues that inform downstream decision making (\cite{tang2026hermes, huang2025vlm, gan2025planning}). This progress has stimulated rapid development of driving-oriented vision--language datasets and benchmarks, which play a central role in making capability claims comparable and reproducible across model families and evaluation settings.

Early driving visual question answering (VQA) and vision--language benchmarks largely follow general-purpose VQA conventions, focusing on scene comprehension, object and relation understanding, and basic spatial reasoning from an ego-centric perspective. Representative datasets emphasize driver behavior explanations and scene narration, such as BDD-X (\cite{kim2018textual}) and DRAMA (\cite{malla2023drama}), while subsequent benchmarks expand question diversity and spatial reasoning coverage, including NuScenes-QA (\cite{qian2024nuscenes}) and NuScenes-SpatialQA (\cite{tian2025nuscenes}). These resources provide an important foundation for evaluating language-based understanding in realistic driving imagery. However, their task formulations and evaluation protocols often remain generic in the sense that they do not explicitly target safety-critical requirements, such as risk-aware reasoning and decision-relevant interpretation under dynamic and uncertain traffic interactions. As a consequence, strong performance on these benchmarks does not necessarily translate to reliable competence on safety-relevant queries that matter for autonomous driving deployment.

Motivated by this gap, recent work has designed benchmarks that more directly emphasize safety, risk, and planning relevance. NuPlanQA (\cite{park2025nuplanqa}) and NuRisk (\cite{gao2025nurisk}) strengthen evaluation by incorporating planning-aware and risk-aware questions, and datasets such as VRU-Accident (\cite{kim2025vru}) highlight safety-relevant edge cases involving vulnerable road users. Complementary evaluation suites, including DriveBench (\cite{kharlamova2025llm}), DVBench (\cite{zeng2025vision}), and ScenePilot-Bench (\cite{wang2026scenepilot}), further stress-test models with more challenging settings and diagnostic protocols, aiming to expose brittleness and failure modes that are masked by easier question distributions. These efforts deepen the evaluation focus and better align driving VQA with safety-oriented objectives. Nevertheless, most of them still assume a single-vehicle, ego-only observability regime. In practice, many safety-critical factors are governed by evidence that is partially visible, occluded, or only apparent at a broader spatiotemporal scale, which limits the extent to which ego-view benchmarks can probe long-range risk buildup, intersection-wide interactions, and macroscopic traffic context.

In parallel, the vehicle-to-everything (V2X) community has developed a rich ecosystem of cooperative perception datasets that explicitly address observability limitations through connectivity and infrastructure sensing. Datasets such as V2X-Sim (\cite{li2022v2x}), V2V4Real (\cite{xu2023v2v4real}), DAIR-V2X (\cite{dair-v2x}), V2X-Seq (\cite{v2x-seq}), V2XReal (\cite{xiang2024v2x}), TUMTraf-V2X (\cite{zimmer2024tumtraf}), UrbanIng-V2X (\cite{sekaran2025urbaning}), and CATS-V2V (\cite{li2025cats}) provide vehicle-side and roadside sensing streams together with synchronization and connectivity metadata, enabling research on cooperative detection, tracking, and related perception tasks. While these resources substantially expand what can be observed and can mitigate occlusion in many cases, they are not designed for language-based evaluation. In particular, they typically lack high-quality question--answer annotations and standardized benchmarking protocols needed to evaluate MLLMs on decision-relevant reasoning grounded in multi-view evidence.

Recent cooperative question answering benchmarks have begun to connect language evaluation with V2X sensing, with examples including V2V-QA (\cite{chiu2025v2v}) and V2V-GoT-QA (\cite{chiu2025v2v-GoT}). These efforts represent an important step toward cooperative understanding, but they remain limited in several respects. First, they primarily focus on vehicle-to-vehicle (V2V) cooperation and do not treat infrastructure sensing as a first-class view. Without a roadside global perspective, cooperative evaluation may resolve certain localized occlusions but cannot consistently provide intersection-level, no-blind-spot macroscopic context. Second, existing setups often evaluate only the cooperative condition and do not support view-decoupled benchmarking that cleanly separates ego-only, infrastructure-only, and cooperative evidence conditions. This limits controlled attribution of performance gains and reduces practical relevance in scenarios where communication is intermittent or unavailable, in which case a model must degrade gracefully to single-view reasoning.

To address these limitations, we introduce V2X-QA, a real-world dataset and benchmark for evaluating MLLMs in autonomous driving across three viewpoints: ego view, infrastructure view from roadside units, and a cooperative setting that jointly uses both views. V2X-QA is designed for view-decoupled evaluation, enabling controlled testing with ego-only, infrastructure-only, and cooperative input under a unified protocol. The benchmark is organized into twelve viewpoint-aligned tasks, including four vehicle-side tasks, four infrastructure-side tasks, and four cooperative tasks, spanning perception, prediction, and reasoning and planning. To standardize both annotation and evaluation, we formulate V2X-QA in a multiple-choice question answering (MCQA) format and construct a task bank covering all twelve tasks, with three questions for each task, together with candidate options and canonical answers for each question. Figure~\ref{fig:v2xqa_overview} presents representative examples from the three viewpoint groups and illustrates how the twelve-task design interfaces with the MCQA-based training and evaluation pipeline of V2X-QA. Within each task, the three questions are assigned to different subsets of visual samples in proportion, so that the question set is used in a balanced manner across the dataset. Based on the corresponding image evidence, a designated multimodal model first selects answers for these task-specific multiple-choice questions, after which the resulting annotations are carefully checked and revised by human experts. The verified data are then split for benchmark construction, and all evaluated models are compared under a unified MCQA prompting protocol. To support controlled studies under this benchmark, we introduce V2X-MoE as a reproducible baseline built upon Qwen3-VL (\cite{bai2025qwen3vl}) with explicit view routing and view-specialized LoRA (\cite{lora_iclr2022}) experts, trained directly for multiple-choice option prediction and further strengthened through targeted cooperative-view and infrastructure-view refinement. We additionally report calibration-based reliability analysis for V2X-MoE to quantify confidence alignment under different observability regimes (\cite{guo2017calibration}). Our contributions are summarized as follows:
\begin{itemize}
    \item \textbf{Dataset.} We introduce V2X-QA, a real-world multi-view autonomous driving dataset spanning ego, infrastructure, and cooperative viewpoints. The dataset contains 33{,}216 annotated instances, and is organized into twelve viewpoint-aligned tasks, including four vehicle-side tasks, four infrastructure-side tasks, and four cooperative tasks, which support fine-grained evaluation across perception, prediction, and reasoning and planning.

    \item \textbf{Benchmark.} We establish a view-decoupled and standardized MCQA benchmark for controlled evaluation under ego-only, infrastructure-only, and cooperative evidence conditions. The benchmark enables reproducible comparison across 10 proprietary and open-source MLLMs under a unified prompting and evaluation protocol.

    \item \textbf{Baseline.} We propose V2X-MoE as a reproducible baseline built upon Qwen3-VL with explicit view routing and view-specialized LoRA experts. It provides a strong and reproducible baseline for viewpoint-aware multiple-choice reasoning under the unified benchmark setting, and we further report calibration-based reliability analysis to quantify confidence alignment across different observability regimes.
\end{itemize}

\section{Related work}
\label{sec:related_work}

\subsection{Ego-view Language benchmarks for autonomous driving}
\label{sec:rw_egoview}

A substantial body of work has established language-based evaluation for autonomous driving under an ego-view setting, where questions are grounded in what the onboard platform can directly observe. Early efforts such as BDD-X (\cite{kim2018textual}) frame driving understanding through textual justification and explanation, linking visual scenes with natural-language rationales for driver behavior. DRAMA (\cite{malla2023drama}) extends this direction by emphasizing richer multimodal grounding for driver attention, action understanding, and scene-aware reasoning. These datasets are important because they introduce the basic paradigm that driving scenes can be evaluated not only through structured outputs, but also through language-mediated interpretation.

Subsequent work moves from explanation-oriented supervision toward more explicit visual question answering. NuScenes-QA (\cite{qian2024nuscenes}) formulates question answering over large-scale autonomous driving data and broadens the range of scene-level semantic queries that models must answer from ego-view evidence. NuScenes-SpatialQA (\cite{tian2025nuscenes}) further emphasizes spatial structure and geometric reasoning, pushing models beyond object naming toward more relation-aware and layout-aware understanding. Together, these works make ego-view driving VQA more systematic and scalable, and they provide a foundation for studying multimodal reasoning in realistic driving imagery.

More recent benchmarks bring the evaluation objective closer to safety-critical deployment. NuPlanQA (\cite{park2025nuplanqa}) introduces planning-oriented question answering and ties language reasoning more directly to downstream driving decisions. NuRisk (\cite{gao2025nurisk}) emphasizes risk-aware understanding, encouraging models to recognize and interpret safety-relevant factors rather than only describe visible content. VRU-Accident (\cite{kim2025vru}) highlights scenarios involving vulnerable road users and accident-related edge cases, thereby making the evaluation distribution more safety-sensitive. In parallel, DriveBench (\cite{kharlamova2025llm}) focuses on diagnostic stress testing, DVBench (\cite{zeng2025vision}) examines more demanding visual reasoning behavior, and ScenePilot-Bench (\cite{wang2026scenepilot}) further enlarges the benchmark scope for driving-oriented multimodal evaluation. Considered together, these datasets show a clear evolution from generic scene description toward planning relevance, risk awareness, and stronger diagnosis of model failure modes.

Despite their differences, these vehicle-centric resources share the same underlying evidence regime: the question is answered from ego-view observations alone. This design is natural for many onboard reasoning problems, but it also imposes a common limitation. Occluded participants, long-range traffic evolution, and intersection-level context may not be fully recoverable from a single vehicle perspective, even when the language question itself is well posed. Consequently, these benchmarks are highly valuable for ego-view multimodal reasoning, yet they do not allow viewpoint accessibility itself to be treated as an experimental variable. From the perspective of benchmark design, this is the key gap left by the vehicle-centric line of work.

\subsection{Cooperative perception corpora}
\label{sec:rw_cooperative_multiview}

In parallel with language-based driving benchmarks, the cooperative autonomous driving community has developed a large family of datasets aimed at overcoming the observability limits of ego-only sensing (\cite{yu2022review}). V2X-Sim (\cite{li2022v2x}) provides an early benchmark for collaborative perception in a simulated setting and helps establish the value of shared multi-agent evidence. DAIR-V2X (\cite{dair-v2x}) marks an important step toward real-world vehicle--infrastructure cooperation by providing synchronized roadside and vehicle-side sensing for cooperative perception tasks. V2V4Real (\cite{xu2023v2v4real}) emphasizes real-world vehicle-to-vehicle collaboration, while V2X-Seq (\cite{v2x-seq}) extends the cooperative setting to sequential perception and forecasting, making temporal context more explicit. More recent datasets, including V2XReal (\cite{xiang2024v2x}), TUMTraf-V2X (\cite{zimmer2024tumtraf}), UrbanIng-V2X (\cite{sekaran2025urbaning}), and CATS-V2V (\cite{li2025cats}), continue to broaden the diversity of sensing configurations, traffic scenarios, and collaboration modes. This line of work has made cooperative perception a mature and technically rich area by demonstrating how connectivity and infrastructure sensing can expand what is observable in real traffic scenes.

However, these datasets are primarily designed for structured perception rather than language-based reasoning. Their main tasks focus on detection, tracking, forecasting, and related predictive outputs, which are essential for cooperative driving but conceptually different from evaluating MLLMs through question answering. As a result, they expand the available evidence for cooperative sensing, but do not provide a benchmark for multi-view reasoning, safety-aware interpretation, and planning-relevant understanding in natural language. Early cooperative QA benchmarks begin to close this gap. V2V-QA (\cite{chiu2025v2v}) demonstrates that question answering can be extended from ego-view driving to a cooperative vehicle-to-vehicle setting, thereby showing that multi-view language reasoning is both feasible and meaningful. V2V-GoT-QA (\cite{chiu2025v2v-GoT}) pushes this direction further by introducing a more structured reasoning perspective for cooperative question answering. These efforts are important because they shift cooperative autonomous driving from purely structured prediction toward language-based evaluation.

At the same time, they still leave two benchmark-level limitations unresolved. First, their cooperative setting is centered on V2V interaction, so infrastructure sensing is not treated as an explicit first-class viewpoint with its own independent reasoning value. Second, they do not provide a fully view-decoupled protocol that evaluates vehicle-only, infrastructure-only, and cooperative conditions under the same task interface. As a result, they do not yet support controlled comparison among different observability regimes. V2X-QA is designed to extend this emerging direction by combining real-world cooperative data, infrastructure participation, language-based question answering, and explicit viewpoint decoupling within a unified benchmark.

\section{Methodology}
\label{sec:method}

\subsection{V2X-QA dataset: views and task taxonomy}
\label{sec:method_dataset}

\subsubsection{View-decoupled problem setup}
\label{sec:view_decoupled_setup}

V2X-QA is designed to evaluate language-based driving reasoning under heterogeneous observability. Each data sample is associated with three viewpoint configurations: a vehicle-side view (VS) captured from the ego platform, an infrastructure-side view (IS) captured from roadside units, and a cooperative view (CO) that jointly leverages both VS and IS evidence. Rather than treating multi-view inputs as an implicit upgrade, V2X-QA makes viewpoint accessibility explicit through a view-decoupled formulation. Specifically, we evaluate models under three evidence conditions corresponding to vehicle-only, infrastructure-only, and cooperative input, while keeping the task interface and benchmark protocol consistent. This design enables controlled attribution of performance differences, separating failures caused by missing evidence from those caused by deficient grounding or reasoning. It also reflects practical operating regimes in which communication or infrastructure availability may vary, requiring models to remain comparable across evidence conditions.

\begin{table}[t]
\centering
\caption{Comparison of driving QA benchmarks, V2X cooperative perception datasets, and cooperative QA resources.
``Multi-perspective decoupling flexibility'' indicates whether the benchmark supports controlled evaluation under different viewpoint accessibility conditions within a consistent protocol.}
\label{tab:dataset_comparison}

\scriptsize
\setlength{\tabcolsep}{4pt}
\renewcommand{\arraystretch}{1.12}

\resizebox{\textwidth}{!}{%
\begin{tabular}{p{0.15\textwidth} p{0.16\textwidth} c c c c c c c c}
\toprule
\makecell[l]{Category} &
Dataset &
Year &
Amount &
\makecell[c]{Connectivity\\\& Cooperation} &
Infrastructure &
\makecell[c]{Multi-Perspective\\Decoupling Flexibility} &
\makecell[c]{Language\\QA} &
\makecell[c]{Safety-Critical\\Reasoning} &
\makecell[c]{Macroscopic\\Traffic Insight} \\
\midrule

\multirow{13}{*}{\makecell[l]{I.\ Single-Vehicle QA\\\textit{("Smart but Blind")}}} &
\makecell[l]{BDD-X\\(\cite{kim2018textual})} & 2018 & 26K   & \xmark & \xmark & \dashmark & \cmark & \xmark & \xmark \\
& \makecell[l]{DRAMA\\(\cite{malla2023drama})} & 2022 & 14K  & \xmark & \xmark & \dashmark & \cmark & \cmark & \xmark \\
& \makecell[l]{NuScenes-SpatialQA\\(\cite{tian2025nuscenes})} & 2022 & 3.3M+ & \xmark & \xmark & \dashmark & \cmark & \xmark & \xmark \\
& \makecell[l]{VRU-Accident\\(\cite{kim2025vru})} & 2023 & 6K & \xmark & \xmark & \dashmark & \cmark & \cmark & \xmark \\
& \makecell[l]{NuScenes-QA\\(\cite{qian2024nuscenes})} & 2024 & 460K & \xmark & \xmark & \dashmark & \cmark & \cmark & \xmark \\
& \makecell[l]{NuPlanQA\\(\cite{park2025nuplanqa})} & 2024 & 1M & \xmark & \xmark & \dashmark & \cmark & \cmark & \xmark \\
& \makecell[l]{NuRisk\\(\cite{gao2025nurisk})} & 2024 & 1.1M & \xmark & \xmark & \dashmark & \cmark & \cmark & \xmark \\
& \makecell[l]{DriveBench\\(\cite{kharlamova2025llm})} & 2024 & 20K & \xmark & \xmark & \dashmark & \cmark & \cmark & \xmark \\
& \makecell[l]{DriveLM\\(\cite{sima2024drivelm})} & 2024 & 443K & \xmark & \xmark & \dashmark & \cmark & \cmark & \xmark \\
& \makecell[l]{DriveMLM\\{\footnotesize (\tabciteDriveMLM)}} & 2024 & 443K & \xmark & \xmark & \dashmark & \cmark & \cmark & \xmark \\
& \makecell[l]{DVBench\\(\cite{zeng2025vision})} & 2025 & 10K & \xmark & \xmark & \dashmark & \cmark & \cmark & \xmark \\
& \makecell[l]{WaymoQA\\{\footnotesize (\tabciteWaymoQA)}} & 2025 & 35K & \xmark & \xmark & \dashmark & \cmark & \cmark & \xmark \\
& \makecell[l]{ScenePilot-4K\\(\cite{wang2026scenepilot})} & 2026 & 27.7M & \xmark & \xmark & \dashmark & \cmark & \cmark & \xmark \\

\midrule

\multirow{8}{*}{\makecell[l]{II.\ V2X Perception\\\textit{("All-Seeing but Mute")}}} &
\makecell[l]{V2X-Sim\\(\cite{li2022v2x})} & 2021 & \dashmark & \cmark & \cmark & \cmark & \xmark & \xmark & \xmark \\
& \makecell[l]{DAIR-V2X\\(\cite{dair-v2x})} & 2022 & \dashmark & \cmark & \cmark & \cmark & \xmark & \xmark & \xmark \\
& \makecell[l]{V2V4Real\\(\cite{xu2023v2v4real})} & 2023 & \dashmark & \cmark & \xmark & \cmark & \xmark & \xmark & \xmark \\
& \makecell[l]{V2X-Seq\\(\cite{v2x-seq})} & 2023 & \dashmark & \cmark & \cmark & \cmark & \xmark & \xmark & \xmark \\
& \makecell[l]{TUMTraf-V2X\\(\cite{zimmer2024tumtraf})} & 2024 & \dashmark & \cmark & \cmark & \cmark & \xmark & \xmark & \xmark \\
& \makecell[l]{V2XReal\\(\cite{xiang2024v2x})} & 2024 & \dashmark & \cmark & \cmark & \cmark & \xmark & \xmark & \xmark \\
& \makecell[l]{UrbanIng-V2X\\(\cite{sekaran2025urbaning})} & 2025 & \dashmark & \cmark & \cmark & \cmark & \xmark & \xmark & \xmark \\
& \makecell[l]{CATS-V2V\\(\cite{li2025cats})} & 2025 & \dashmark & \cmark & \xmark & \cmark & \xmark & \xmark & \xmark \\

\midrule

\multirow{3}{*}{\makecell[l]{III.\ Cooperative QA}} &
\makecell[l]{V2V-QA\\(\cite{chiu2025v2v})} & 2025 & 1.45M & \cmark & \xmark & \xmark & \cmark & \cmark & \xmark \\
& \makecell[l]{V2V-GoT-QA\\(\cite{chiu2025v2v-GoT})} & 2025 & 141K & \cmark & \xmark & \xmark & \cmark & \cmark & \xmark \\
& \makecell[l]{V2X-QA (Ours)} & 2026 & 33.2K & \cmark & \cmark & \cmark & \cmark & \cmark & \cmark \\

\bottomrule
\end{tabular}
}
\end{table}

Table~\ref{tab:dataset_comparison} summarizes how existing resources relate to this objective. Driving QA benchmarks provide language supervision and reasoning-oriented evaluation but typically lack infrastructure sensing and controlled degradation to infrastructure-only evidence. V2X cooperative perception datasets provide connectivity and multi-perspective sensing but are not designed for language-based QA and standardized reasoning evaluation. Recent cooperative QA resources begin to connect multi-perspective sensing with QA, yet they often do not treat infrastructure as a first-class view and do not support principled fallback to single-view evidence conditions under the same protocol. These gaps motivate a unified dataset and benchmark that explicitly models viewpoint accessibility and enables controlled and standardized comparisons across comprehensive VS, IS, and CO settings.

\nocite{cui2025drivemlm}
\nocite{yu2025waymoqa}

\subsubsection{Twelve-task taxonomy}
\label{sec:task_taxonomy}

V2X-QA is organized around a twelve-task taxonomy that serves as the core semantic structure of the dataset and benchmark. Each sample is assigned exactly one task label, so the taxonomy is not an auxiliary categorization after data construction, but an integral part of how V2X-QA is annotated, organized, and evaluated. The twelve tasks are grouped into three viewpoint-aligned sets, corresponding to vehicle-side understanding, infrastructure-side macroscopic understanding, and cooperative cross-view reasoning. Across these viewpoint-aligned groups, the tasks cover three capability dimensions that are central to autonomous driving question answering: perception-oriented scene understanding, short-horizon behavior anticipation, and safety-aware reasoning and planning.

For vehicle-side understanding, the taxonomy includes four tasks: \textit{Ego Static Scene} (VS1), \textit{Ego Visible Agents} (VS2), \textit{Ego Risk} (VS3), and \textit{Ego Prediction} (VS4). These tasks focus on evidence naturally available from the ego perspective, including static road structure and traffic control cues, nearby traffic participants and their relative configurations, localized safety risks, and short-horizon anticipation of salient agents or ego-relevant motion.

For infrastructure-side macroscopic understanding, the taxonomy includes four tasks: \textit{RSU Global Layout} (IS1), \textit{RSU Traffic Agents} (IS2), \textit{RSU Global Risk} (IS3), and \textit{RSU Long-Range Cues} (IS4). This group leverages the roadside vantage and long-range coverage to emphasize global layout and topology, aggregated agent configuration over a wider field of view, macroscopic risk factors that may be weak or occluded in the ego view, and far-field cues enabled by infrastructure sensing. Importantly, the infrastructure-only setting supports intersection-level and corridor-level analysis, where a roadside view can summarize global traffic state and safety-relevant context even without ego observations.

For cooperative cross-view reasoning, the taxonomy includes four tasks: \textit{Cooperative Scene Understanding} (CO1), \textit{Cooperative Visibility} (CO2), \textit{Cooperative Prediction} (CO3), and \textit{Cooperative Planning} (CO4). This group targets evidence integration across viewpoints, including cross-view grounding and reconciliation of partial observations, explicit reasoning about visibility and occlusion, improved short-horizon prediction using complementary cues, and planning-relevant decisions grounded in jointly observed context.

This taxonomy enables fine-grained diagnosis that is aligned with viewpoint accessibility. When a model fails under vehicle-only evidence but succeeds under cooperative evidence for the same task family, the failure is likely dominated by missing cues rather than purely linguistic or logical deficits. Conversely, consistent failures across evidence conditions indicate intrinsic limitations in grounding or reasoning. 

\subsection{Data curation and benchmark protocol}
\label{sec:data_curation_benchmark}

\subsubsection{MCQA-native annotation with human verification}
\label{sec:annotation_verification}

\begin{figure}[pos=htbp]
    \centering
    \includegraphics[width=\textwidth]{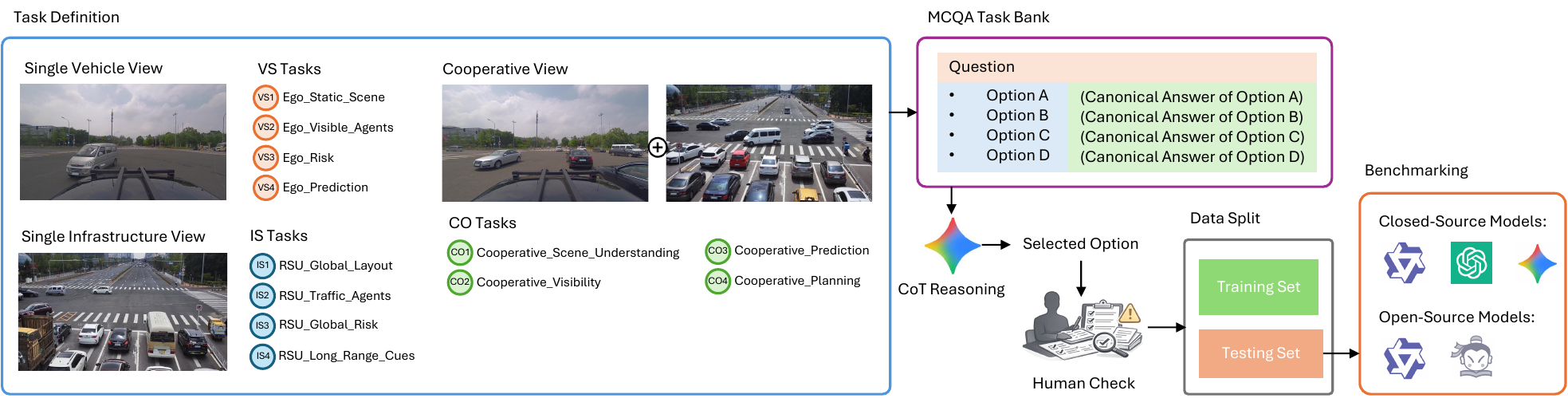}
    \caption{Construction pipeline of V2X-QA. The pipeline starts from the twelve viewpoint-aligned tasks under three settings, then proceeds to MCQA task bank construction, model-assisted answer selection with human verification, data split generation, and standardized benchmarking across proprietary and open-source MLLMs.}
    \label{fig:annotation_pipeline}
\end{figure}

V2X-QA is built on real-world vehicle--infrastructure cooperative driving data derived from the V2X-Seq line of datasets (\cite{v2x-seq}). We use the V2X-Seq-SPD setting as the underlying source domain, which extends the sequential V2X benchmark with synchronized vehicle-side and roadside observations for cooperative driving and has been adopted in recent end-to-end V2X evaluation pipelines (\cite{you2026v2x, you2025seal}). For dataset construction, we sample 6,000 vehicle-side images from V2X-Seq and retain the corresponding matched infrastructure-side views when available. This yields 6,000 vehicle-side scenes and 5,304 synchronized vehicle--infrastructure pairs for infrastructure-side and cooperative annotation, allowing the same scene context to be queried under vehicle-only, infrastructure-only, and cooperative evidence conditions.

Based on the twelve-task taxonomy defined in Section~\ref{sec:task_taxonomy}, we formulate V2X-QA in an MCQA format. Specifically, we construct an MCQA task bank covering all twelve tasks, where each task is associated with three predefined questions, candidate options, and canonical answers. Within each task, the three questions are assigned to different subsets of samples in roughly balanced proportions, so that the question set is used in a consistent and well-covered manner across the dataset. The complete task bank is provided in Appendix~\ref{app:mcqa_task_bank}.

For each sample, we use Gemini-2.5-Pro (\cite{comanici2025gemini25}) to select an answer option for the assigned task-specific question based on the corresponding image evidence. The resulting annotations are then carefully checked and revised by human experts. The review process verifies visual grounding, correctness of the selected option, consistency of the canonical answer, and compliance with the intended task and viewpoint setting. Samples that fail these criteria are corrected. After verification, the curated data are partitioned into training and testing sets using a question-aware and pair-consistent split strategy. Specifically, we first select the test subset from the cooperative subset by sampling 10\% of synchronized vehicle--infrastructure pairs within each question group, so that all predefined questions remain represented in the test set in a balanced manner. The selected cooperative test pairs are then projected to their corresponding vehicle-side and infrastructure-side image IDs, and the same projected IDs are used to split the VS and IS subsets. In this way, the train or test partition preserves cross-view consistency and avoids leakage caused by matched vehicle-side, infrastructure-side, or cooperative samples appearing across both splits. Figure~\ref{fig:annotation_pipeline} summarizes the full workflow from task definition and MCQA construction to human verification, data split generation, and final benchmarking.

\subsubsection{Dataset statistics and answer-distribution diagnostics}
\label{sec:mcqa_construction}

\begin{figure}[pos=htbp]
    \centering
    \includegraphics[width=\textwidth]{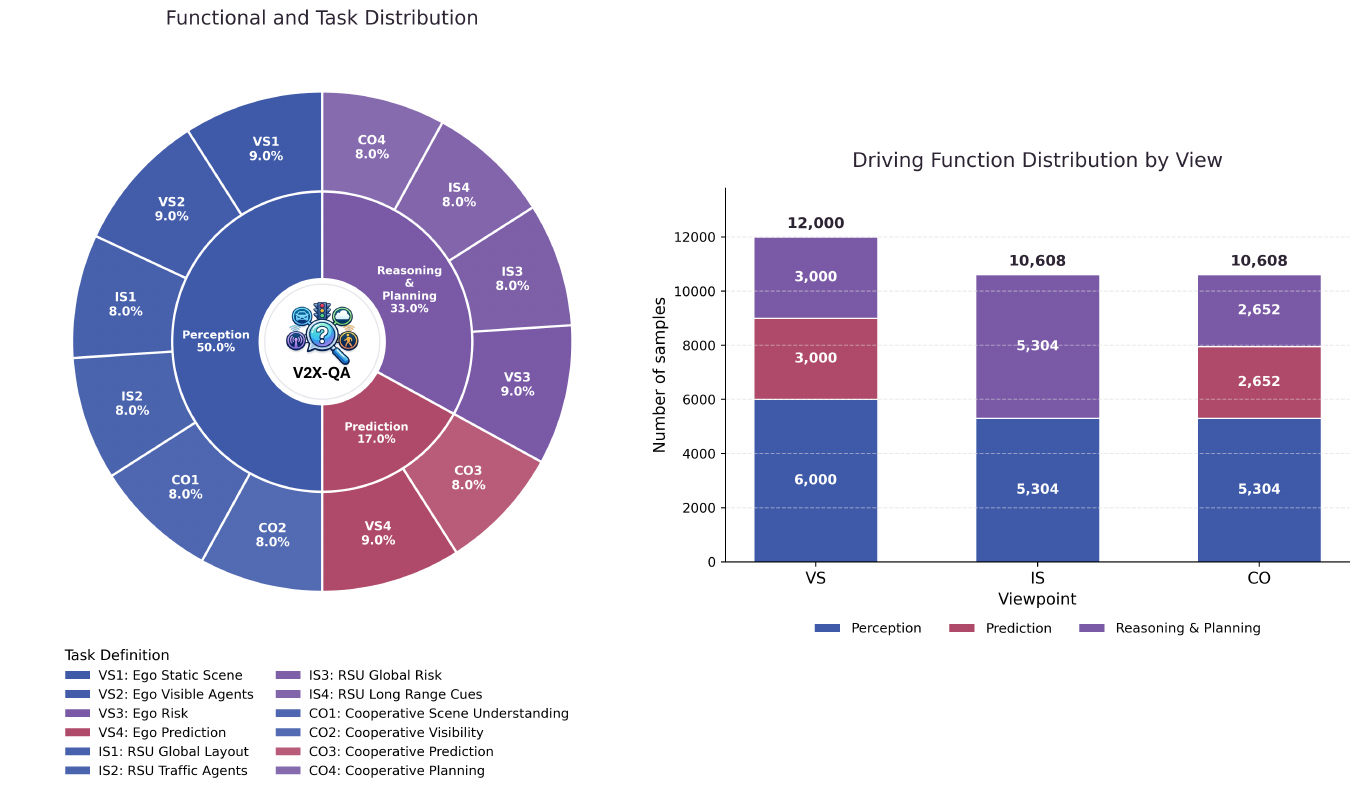}
    \caption{Statistics of V2X-QA. Left: overall task and functional distribution across the twelve viewpoint-aligned tasks. Right: distribution of perception, prediction, and reasoning \& planning samples under vehicle-side, infrastructure-side, and cooperative views.}
    \label{fig:dataset_statistics}
\end{figure}

\begin{figure}[pos=htbp]
    \centering
    \includegraphics[width=\textwidth]{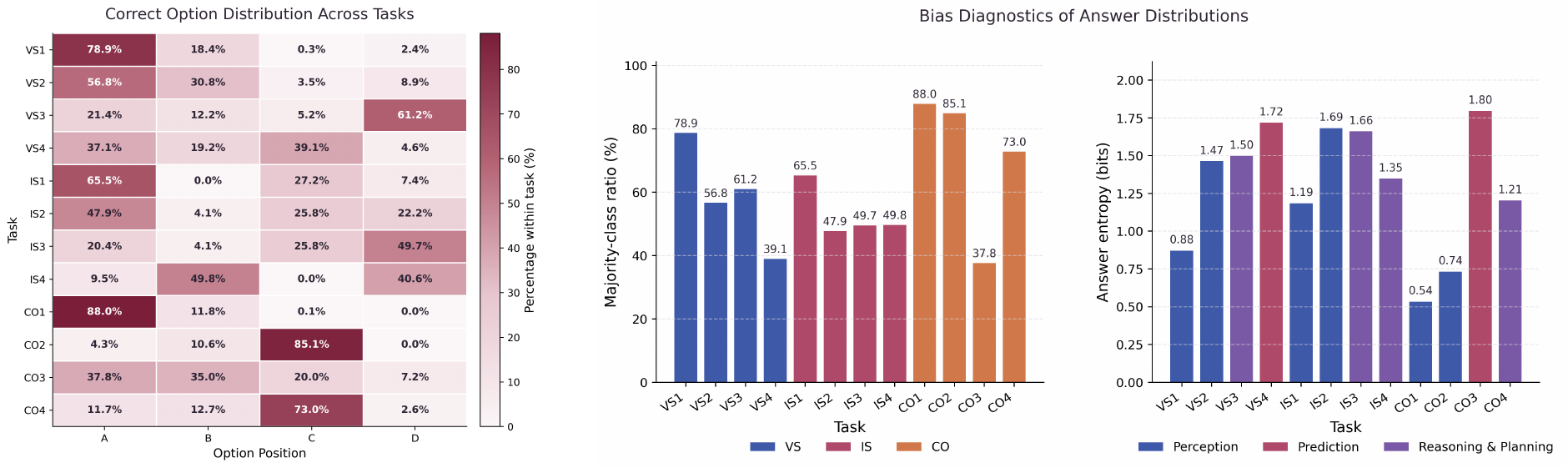}
    \caption{Answer-distribution diagnostics of V2X-QA. Left: distribution of correct option positions across the twelve tasks. Middle: majority-class ratio for each task. Right: answer entropy for each task.}
    \label{fig:answer_bias}
\end{figure}

Figure~\ref{fig:dataset_statistics} summarizes the resulting data distribution. The vehicle-side subset contains 12{,}000 samples in total, corresponding to 3{,}000 samples for each of the four VS tasks. Because matched infrastructure-side views are available for a subset of the sampled vehicle-side images, the infrastructure-side subset contains 10{,}608 samples in total, corresponding to 2{,}652 samples for each of the four IS tasks. The cooperative subset follows the same matched-pair constraint and likewise contains 10{,}608 samples, again with 2{,}652 samples for each of the four CO tasks. Aggregated by functional category, the full benchmark consists of 50.0\% perception samples, 17.0\% prediction samples, and 33.0\% reasoning and planning samples. View-wise composition is also consistent with the task definitions: the VS subset includes perception, prediction, and reasoning and planning samples; the IS subset emphasizes perception together with macroscopic reasoning; and the CO subset covers perception, cooperative prediction, and cooperative planning.

Since V2X-QA adopts a fixed-option MCQA formulation, it is also important to characterize answer-distribution bias. Figure~\ref{fig:answer_bias} reports the distribution of correct option positions across tasks together with two complementary diagnostics: the majority-class ratio and the answer entropy. These statistics show that the option distribution is task-dependent rather than uniformly balanced. This pattern is expected in real-world driving data, because normal traffic scenes are often dominated by common and repeated situations, whereas rarer behaviors, unusual interactions, and long-tail scenarios appear much less frequently. As a result, some tasks naturally exhibit a stronger concentration on particular options, while others show a more diverse answer distribution.

\subsection{Baseline: V2X-MoE and evaluation setup}
\label{sec:v2xmoe_setup}

\subsubsection{Architecture of V2X-MoE and training and evaluation pipeline}
\label{sec:v2xmoe_architecture}

\begin{figure}[pos=htbp]
    \centering
    \includegraphics[width=\textwidth]{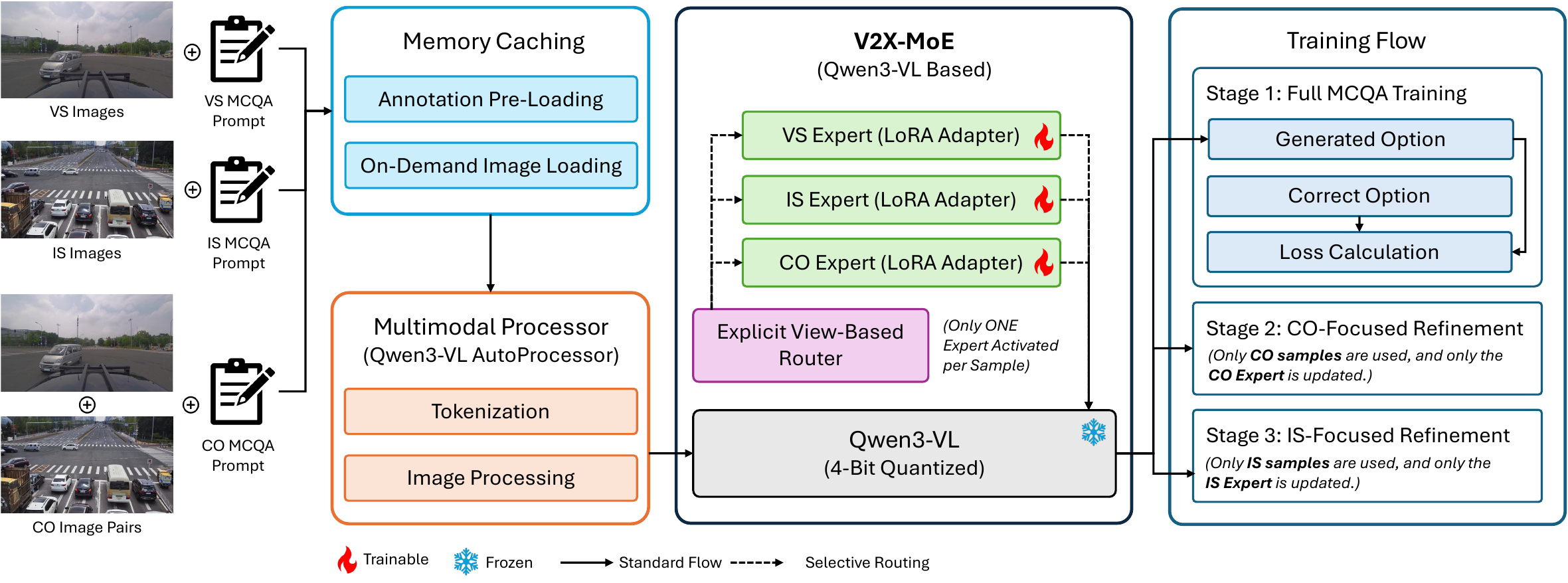}
    \caption{Overall architecture and staged training pipeline of V2X-MoE. The model takes viewpoint-aligned visual evidence and MCQA prompts as input, uses a Qwen3-VL multimodal processor and a shared frozen backbone, activates one viewpoint-specific LoRA expert through an explicit router, and is trained by full MCQA training followed by CO-focused and IS-focused refinement.}
    \label{fig:v2xmoe_system}
\end{figure}

To support controlled benchmarking under the view-decoupled setting, we introduce V2X-MoE as a reproducible baseline built upon Qwen3-VL (\cite{bai2025qwen3vl}). The design follows a simple principle: a shared multimodal backbone is used across all samples, while viewpoint-specific adaptation is introduced through lightweight experts that are explicitly selected according to the evidence condition. Given an input sample, the model receives a viewpoint type signal $v \in \{\mathrm{VS}, \mathrm{IS}, \mathrm{CO}\}$ together with the corresponding visual evidence and MCQA prompt. Depending on the benchmark condition defined in Section~\ref{sec:view_decoupled_setup}, the visual input is either a vehicle-side image, an infrastructure-side image, or a paired cooperative input. For cooperative samples, the image order is fixed as vehicle-side first and infrastructure-side second. After multimodal preprocessing with the Qwen3-VL AutoProcessor, the resulting token sequence is passed through a frozen Qwen3-VL backbone, while a viewpoint router activates exactly one expert associated with the current evidence condition.

Formally, let $x$ denote the tokenized multimodal input and let $f_{\theta}$ denote the frozen Qwen3-VL backbone parameterized by $\theta$. We introduce three lightweight expert modules, denoted as $\mathcal{E}_{\mathrm{VS}}$, $\mathcal{E}_{\mathrm{IS}}$, and $\mathcal{E}_{\mathrm{CO}}$, together with a deterministic router
\begin{equation}
r(v) \in \{\mathcal{E}_{\mathrm{VS}}, \mathcal{E}_{\mathrm{IS}}, \mathcal{E}_{\mathrm{CO}}\},
\end{equation}
which selects the expert that matches the viewpoint condition $v$. The resulting model is therefore conditioned on both the input and the selected expert
\begin{equation}
f_{\Theta}(x,v)=f_{\theta,\phi_{r(v)}}(x),
\end{equation}
where $\phi_{r(v)}$ denotes the trainable parameters of the selected expert and
\begin{equation}
\Theta = \{\theta, \phi_{\mathrm{VS}}, \phi_{\mathrm{IS}}, \phi_{\mathrm{CO}}\}.
\end{equation}
In our implementation, the shared backbone remains frozen and only the expert parameters are optimized. This design keeps the baseline lightweight while making the role of viewpoint specialization explicit and interpretable.

V2X-MoE is trained directly for MCQA answer generation, where the target output is the correct option letter itself. For a training sample $(x_i, v_i, s_i)$, let
\begin{equation}
s_i=(s_{i,1},\ldots,s_{i,T_i})
\end{equation}
denote the target answer sequence encoding the correct option. In practice, this is a very short sequence consisting of the gold option letter followed by the end-of-sequence token. The training objective is the standard autoregressive negative log-likelihood:
\begin{equation}
\mathcal{L}_{\mathrm{MCQA}}
=
-\sum_{i=1}^{N}\sum_{t=1}^{T_i}
\log p_{\Theta}\!\left(s_{i,t}\mid x_i, v_i, s_{i,<t}\right).
\end{equation}
In this way, V2X-MoE preserves the causal language modeling interface of Qwen3-VL while aligning supervision directly with the benchmark answer format.

To reduce spurious bias toward fixed option positions, the option order is shuffled during training and the gold label is remapped accordingly. We further apply task-balanced sampling during optimization to stabilize learning across the twelve tasks. On top of this general training stage, we adopt two targeted refinement stages. First, we perform CO-focused refinement using only cooperative samples, during which only the cooperative expert is updated and the prompt explicitly encourages joint use of the two views. Second, we perform IS-focused refinement using only infrastructure-side samples, during which only the infrastructure expert is updated and the prompt emphasizes global layout, traffic participants, and long-range roadside cues. This stage-wise design improves viewpoint specialization while preserving the shared backbone and the explicit routing structure.

During evaluation, we follow the same prompt-based MCQA protocol used by the benchmark and obtain the model's answer through generation. For an input sample $(x,v)$, the model generates a short response
\begin{equation}
\tilde{s}=\mathrm{Generate}(x,v),
\end{equation}
which is then parsed into a valid option label
\begin{equation}
\hat{y}=\mathrm{Parse}(\tilde{s}), \qquad \hat{y}\in\{A,B,C,D\}.
\end{equation}
Accuracy is computed from the resulting parsed option predictions. This evaluation rule matches the actual prompt-and-generation behavior of the model and therefore remains directly comparable with other models benchmarked on V2X-QA.

For calibration analysis, we additionally extract the first-step logits over the four candidate option letters under the same prompt context. Let $z_A, z_B, z_C, z_D$ denote the corresponding logits. The option probabilities are computed as:
\begin{equation}
p_{\Theta}(y \mid x,v)
=
\frac{\exp(z_y)}
{\sum_{y' \in \{A,B,C,D\}} \exp(z_{y'})},
\qquad y\in\{A,B,C,D\},
\end{equation}
and the confidence score is defined as:
\begin{equation}
\hat{p}=\max_{y\in\{A,B,C,D\}} p_{\Theta}(y \mid x,v).
\end{equation}
These probabilities are used only for confidence estimation and reliability analysis; they do not replace the generation-based prediction rule used for benchmark accuracy.

Figure~\ref{fig:v2xmoe_system} illustrates the full V2X-MoE pipeline. The figure highlights three properties that are central to our benchmark setting. First, viewpoint-specific evidence conditions are aligned with viewpoint-specific MCQA prompts at the data interface. Second, the router enforces explicit expert specialization rather than implicit mixture weighting. Third, the training procedure is stage-wise and benchmark-aligned, consisting of full MCQA training followed by targeted refinement for cooperative and infrastructure-side reasoning.

\subsubsection{View-specific LoRA injection with explicit routing}
\label{sec:v2xmoe_lora}

Viewpoint specialization in V2X-MoE is implemented through LoRA-based adaptation (\cite{lora_iclr2022}), applied to the attention projections of the transformer blocks. Specifically, low-rank updates are injected into the query, key, value, and output projection matrices of self-attention, corresponding to the standard projection operators \(W_q\), \(W_k\), \(W_v\), and \(W_o\). Let $x \in \mathbb{R}^{d}$ denote an input token representation at a given layer. In a standard transformer block, the attention projections are
\begin{equation}
q = W_q x,\qquad
k = W_k x,\qquad
v = W_v x,
\end{equation}
and the output projection is
\begin{equation}
o = W_o z,
\end{equation}
where $z$ is the self-attention output and $W_q, W_k, W_v, W_o$ are the base projection matrices. In V2X-MoE, these backbone projection matrices remain frozen, and viewpoint-specific low-rank updates are injected through the selected expert.

For viewpoint condition $v \in \{\mathrm{VS}, \mathrm{IS}, \mathrm{CO}\}$, let the corresponding expert parameters be
\begin{equation}
\Delta W_q^{(v)} = B_q^{(v)}A_q^{(v)}, \quad
\Delta W_k^{(v)} = B_k^{(v)}A_k^{(v)}, \quad
\Delta W_v^{(v)} = B_v^{(v)}A_v^{(v)}, \quad
\Delta W_o^{(v)} = B_o^{(v)}A_o^{(v)},
\end{equation}
where each $A_{\bullet}^{(v)} \in \mathbb{R}^{r \times d}$ and $B_{\bullet}^{(v)} \in \mathbb{R}^{d \times r}$ defines a rank-$r$ LoRA update. The viewpoint-conditioned projections become
\begin{equation}
q^{(v)} = \left(W_q + \Delta W_q^{(v)}\right)x,\qquad
k^{(v)} = \left(W_k + \Delta W_k^{(v)}\right)x,\qquad
v^{(v)} = \left(W_v + \Delta W_v^{(v)}\right)x,
\end{equation}
and
\begin{equation}
o^{(v)} = \left(W_o + \Delta W_o^{(v)}\right)z.
\end{equation}
Because only one expert is active for each sample, these updates are never mixed across viewpoints within a forward pass. The router therefore induces an explicit hard partition over expert usage
\begin{equation}
\Delta W_{\bullet}(v) =
\begin{cases}
\Delta W_{\bullet}^{(\mathrm{VS})}, & v=\mathrm{VS},\\
\Delta W_{\bullet}^{(\mathrm{IS})}, & v=\mathrm{IS},\\
\Delta W_{\bullet}^{(\mathrm{CO})}, & v=\mathrm{CO}.
\end{cases}
\end{equation}
This hard-routing mechanism is consistent with the benchmark design, because each sample is evaluated under one evidence condition at a time. It also makes interpretation straightforward: any adaptation applied by the model can be directly attributed to the corresponding viewpoint expert.

The rationale for this design is twofold. First, it preserves a common multimodal backbone across all samples, which keeps the baseline comparable across evidence conditions and avoids introducing unnecessary model heterogeneity. Second, it allows the trainable capacity to focus on viewpoint-specific deviations from the shared representation. The VS expert can specialize in ego-centric local semantics and immediate risk cues, the IS expert can emphasize global topology and long-range traffic organization, and the CO expert can specialize in cross-view integration and cooperative reasoning. Because these adaptations are introduced at the attention projection level, they influence how the model forms and propagates viewpoint-specific interactions throughout the transformer stack. The stage-wise training procedure further reinforces this specialization by refining the CO and IS experts on their corresponding subsets after full MCQA training.

\begin{figure}[pos=htbp]
    \centering
    \includegraphics[width=0.95\textwidth]{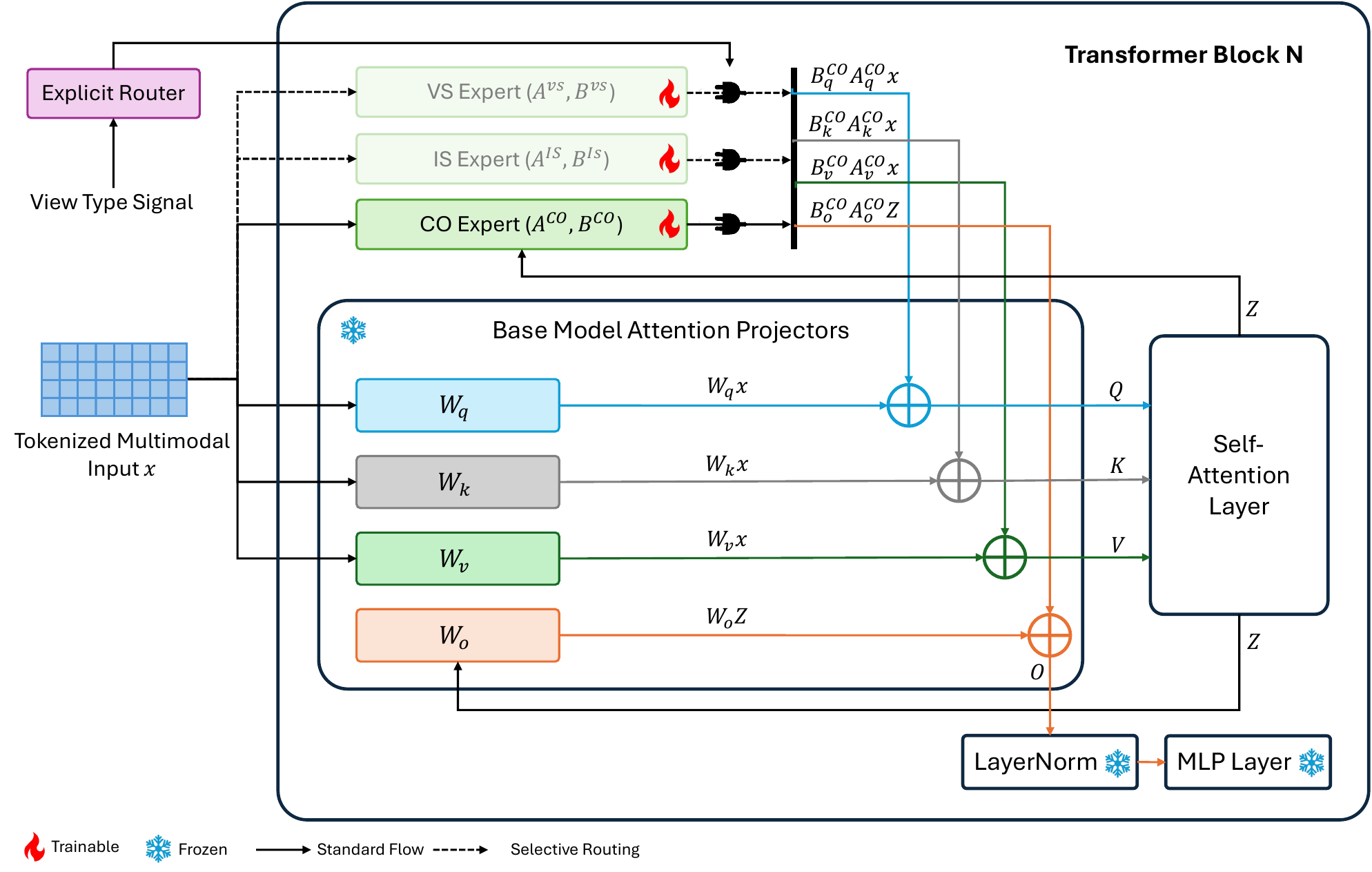}
    \caption{View-specific LoRA injection with explicit hard routing in V2X-MoE. A viewpoint type signal selects one expert among the vehicle-side, infrastructure-side, and cooperative experts. The selected expert injects low-rank adaptations into the attention projections, while the base transformer parameters remain frozen.}
    \label{fig:lora_injection}
\end{figure}

Figure~\ref{fig:lora_injection} visualizes this mechanism. The figure shows how the explicit router selects one viewpoint expert, how the corresponding LoRA parameters are injected into the query, key, value, and output projections, and how the frozen backbone computations are preserved elsewhere. In this way, V2X-MoE remains a lightweight and controlled baseline while still exposing a concrete mechanism for viewpoint specialization that aligns naturally with the task structure of V2X-QA.

In all experiments, we report standardized MCQA accuracy for both proprietary and open-source MLLMs. In addition, for V2X-MoE we report calibration-based reliability metrics to quantify confidence alignment under different evidence conditions. Given the predicted confidence $\hat{p}_i$ for sample $i$ and its correctness indicator $c_i \in \{0,1\}$, Expected Calibration Error (ECE) (\cite{guo2017calibration}) is computed as:
\begin{equation}
\mathrm{ECE}
=
\sum_{m=1}^{M}
\frac{|B_m|}{n}
\left|
\mathrm{acc}(B_m)-\mathrm{conf}(B_m)
\right|,
\end{equation}
where $B_m$ denotes the set of samples falling into confidence bin $m$, $\mathrm{acc}(B_m)$ is the empirical accuracy in that bin, $\mathrm{conf}(B_m)$ is the average confidence, and $n$ is the total number of evaluated samples. We additionally report the Brier score (\citep{brier1950verification}) as a complementary measure of probabilistic prediction quality. Let
\begin{equation}
\mathbf{p}_i =
\bigl[
p_{\Theta}(y=A \mid x_i, v_i),\,
p_{\Theta}(y=B \mid x_i, v_i),\,
p_{\Theta}(y=C \mid x_i, v_i),\,
p_{\Theta}(y=D \mid x_i, v_i)
\bigr]
\end{equation}
denote the predicted four-way option probability vector for sample \(i\), and let $\mathbf{y}_i \in \{0,1\}^{4}$ denote the corresponding one-hot ground-truth label vector. The Brier score is then computed as:
\begin{equation}
\mathrm{Brier}
=
\frac{1}{n}\sum_{i=1}^{n}\sum_{k \in \{A,B,C,D\}}
\bigl(p_{\Theta}(y=k \mid x_i, v_i)-y_{i,k}\bigr)^2.
\end{equation}
In our implementation, this is computed from the first-step option probabilities over the four candidate letters \(A\), \(B\), \(C\), and \(D\) under the same prompt context, by constructing the corresponding one-hot target vector for the correct option and summing the squared differences between the two vectors for each sample. These reliability metrics are used only for V2X-MoE and are intended to complement, rather than replace, the benchmark's main accuracy-based evaluation.

\section{Experiments}
\label{sec:experiments}

\subsection{Experimental setup}
\label{sec:exp_setup}

This section summarizes the common evaluation setting used throughout the benchmark and the implementation details of the V2X-MoE baseline. Unless otherwise noted, all models are evaluated under the same multiple-choice protocol introduced in Section~\ref{sec:mcqa_construction}, and all reported results are obtained on the expert-verified V2X-QA test split.

\subsubsection{Evaluated models}
\label{sec:exp_models}

We evaluate a diverse set of proprietary and open-source MLLMs. The evaluated model pool includes both recent frontier systems and representative open-source VLMs, and the complete model list is reported together with the corresponding result tables in Section~\ref{sec:main_results}. In addition to these general-purpose MLLMs, we include V2X-MoE as a reproducible baseline specifically designed for the view-decoupled setting of V2X-QA. In its current implementation, V2X-MoE is built upon Qwen3-VL-8B-Instruct (\cite{bai2025qwen3vl}) with explicit view routing and viewpoint-specific LoRA experts.

\subsubsection{Evaluation protocol}
\label{sec:exp_protocol}

All benchmark comparisons are conducted under the standardized MCQA setting. For each test sample, a model receives the visual evidence associated with the designated viewpoint condition together with the question and four candidate answers. The prompt explicitly specifies the viewpoint condition and instructs the model to return a single uppercase option letter. To ensure standardized and reproducible benchmarking across all evaluated models, we use a unified benchmark prompt template, which is provided in Appendix~\ref{app:benchmark_prompt}. This unified answer interface ensures that benchmark comparisons are not confounded by free-form response style or differences in answer verbosity.

For proprietary and open-source MLLMs, we evaluate their option predictions directly under this MCQA formulation and compute accuracy over the selected answer. For V2X-MoE, evaluation is also conducted under the same prompt-based MCQA interface. The model generates a short response, which is then parsed into one of the four valid option labels $\{A,B,C,D\}$. Reported benchmark results are further aggregated by viewpoint group, task label, and the full test set.

\subsubsection{Metrics}
\label{sec:exp_metrics}

The primary metric of V2X-QA is MCQA accuracy. We report overall accuracy and fine-grained breakdowns across the three viewpoint groups and the twelve task labels. These results form the main benchmark comparison among proprietary models, open-source models, and the V2X-MoE baseline.

For V2X-MoE only, we additionally report calibration-based reliability analysis, including ECE and Brier score, to quantify confidence alignment under different evidence conditions. These reliability metrics are used as supplementary diagnostics for the baseline and are not treated as benchmark-wide metrics for all evaluated models.

\subsubsection{Implementation details}
\label{sec:exp_impl}

V2X-MoE is implemented on top of Qwen3-VL-8B-Instruct (\cite{bai2025qwen3vl}). The base model is loaded with a frozen multimodal backbone, and viewpoint-specific LoRA adapters are attached to the self-attention projections. The implementation uses 4-bit NF4 quantization with bfloat16 computation and FlashAttention-2 for efficient training and inference. Three LoRA experts are defined for the vehicle-side, infrastructure-side, and cooperative conditions, respectively, and each sample activates exactly one expert according to its viewpoint type.

For all V2X-MoE experts, the LoRA rank $r$ is set to 16, the scaling factor $\alpha$ is 32, and the dropout rate is 0.05. Following the viewpoint-specific adaptation mechanism introduced in Section~\ref{sec:v2xmoe_lora}, we inject low-rank updates into the self-attention projections \(W_q\), \(W_k\), \(W_v\), and \(W_o\), so that the selected expert can modulate how viewpoint-conditioned information is formed and propagated within the transformer blocks. The shared Qwen3-VL backbone and vision tower remain frozen throughout training, and only the LoRA parameters of the activated expert are updated.

V2X-MoE is trained on the V2X-QA training split using MCQA-aligned autoregressive supervision. During training, the prompt includes the question and four candidate options, while the target output is the correct option letter itself. To reduce overfitting to fixed option positions, the option order is shuffled online and the gold label is remapped accordingly. We further apply task-balanced sampling during optimization to reduce task imbalance across the twelve tasks.

The full MCQA training stage is performed for 4 epochs with batch size 1, gradient accumulation over 8 steps, learning rate \(1\times10^{-4}\), weight decay \(0.01\), and warmup ratio \(0.03\). Images are resized with a maximum side length of 560, and the processor maximum is set to \(560\times560\) pixels. After this full training stage, we conduct a CO-focused refinement stage for 2 epochs with learning rate \(5\times10^{-5}\) and warmup ratio \(0.05\), during which only the cooperative expert is updated. We then conduct an IS-focused refinement stage, again for 2 epochs with learning rate \(5\times10^{-5}\) and warmup ratio \(0.05\), during which only the infrastructure expert is updated. In the IS-focused stage, we additionally place slightly higher sampling weight on the more difficult IS3 and IS4 tasks.

During inference, the same viewpoint signal selects the corresponding expert, and evaluation is performed on MCQA samples from the test split. For standard benchmark reporting, the model predicts a single answer option through generation, and accuracy is computed from the parsed option label. For reliability analysis, confidence scores are derived from the first-step probabilities over the candidate option letters under the same prompt context, and ECE is computed using 12 confidence bins. This yields a consistent probabilistic interface for calibration analysis while preserving the same MCQA evaluation protocol used for all models in the benchmark.

\subsection{Main benchmark results}
\label{sec:main_results}

We now evaluate how current MLLMs perform on V2X-QA under the standardized MCQA protocol. The main benchmark results are reported for the vehicle-side and infrastructure-side tasks in Table~\ref{tab:main_vs_is_results} and for the cooperative tasks in Table~\ref{tab:main_co_results}. Taken together, these results answer the central benchmark question of this work: how well do current MLLMs reason under vehicle-side, infrastructure-side, and cooperative observability regimes, and to what extent can task- and viewpoint-specific adaptation improve performance under a unified evaluation setting.

\subsubsection{Vehicle-side and infrastructure-side results}
\label{sec:results_vs_is}

Table~\ref{tab:main_vs_is_results} summarizes the task-wise accuracy on the four vehicle-side tasks and the four infrastructure-side tasks. A first observation is that off-the-shelf MLLMs remain far from saturated on both single-view regimes. Among the proprietary models, Qwen-3.5-Plus achieves the highest vehicle-side average of 60.8, while Qwen3-VL-Flash reaches the highest proprietary infrastructure-side average of 54.7. Among the open-source models, the best vehicle-side performance comes from Intern-S1 with an average of 58.5, whereas the strongest infrastructure-side average is 59.7, jointly achieved by Intern-S1 and Qwen-2.5-VL-72B-Instruct. These results indicate that even under single-view conditions, V2X-QA is not solved by current general-purpose MLLMs.

A second observation is that performance varies substantially across tasks within the same viewpoint group. On the vehicle-side tasks, models are generally stronger on VS1 and VS2 than on VS3 and VS4, suggesting that scene recognition and visible-agent understanding are easier than risk assessment and short-horizon prediction. On the infrastructure-side tasks, several models perform reasonably well on IS1 and IS2 yet remain less stable on IS3 and IS4, which require macroscopic risk interpretation and long-range cue extraction from roadside observations. This pattern confirms that infrastructure-side reasoning is not merely an auxiliary variant of ego-view understanding, but a distinct benchmark regime emphasizing global topology, broader traffic organization, and far-field evidence.

The most important result is that V2X-MoE substantially outperforms all compared models on both viewpoint groups. It achieves an average accuracy of 95.3 on the vehicle-side tasks and 94.0 on the infrastructure-side tasks, while also maintaining uniformly strong performance across all individual tasks. The gain is especially notable on the harder reasoning-oriented tasks, including VS3, VS4, IS3, and IS4, where the proposed viewpoint-specific design remains consistently robust rather than excelling only on a subset of cases. This result suggests that explicit view routing, view-specialized LoRA adaptation, and staged refinement are highly effective for learning viewpoint-conditioned reasoning under the V2X-QA benchmark.

At the same time, this comparison should be interpreted in light of the evaluation setting. The proprietary and open-source MLLMs are benchmarked under the unified prompt protocol without task-specific adaptation, whereas V2X-MoE is trained on the V2X-QA training split. Its strong performance therefore demonstrates not only the difficulty of the benchmark for general-purpose models, but also the effectiveness of benchmark-aligned specialization when viewpoint structure is modeled explicitly.

\begin{table*}[t]
\centering
\caption{Main benchmark results on the vehicle-side and infrastructure-side tasks of V2X-QA. Accuracy (\%) is reported for each task and the corresponding average within each viewpoint group.}
\label{tab:main_vs_is_results}
\setlength{\tabcolsep}{5pt}
\renewcommand{\arraystretch}{1.15}
\small
\begin{tabular}{lccccc|ccccc}
\toprule
Model & VS1 & VS2 & VS3 & VS4 & Avg. & IS1 & IS2 & IS3 & IS4 & Avg. \\
\midrule
\multicolumn{11}{l}{\textit{Closed-Source Models}} \\
GPT-5.2 (\cite{openai2025gpt52})& 65.3 & 43.8 & 26.0 & 43.0 & 44.5 & 69.8 & 42.6 & 38.5 & 47.5 & 49.6 \\
GPT-5-Mini (\cite{openai2025gpt5mini})& 78.1 & 49.8 & 32.1 & 44.5 & 51.1 & 45.7 & 52.1 & 46.4 & 45.3 & 47.4 \\
Gemini-3-Flash-Preview (\cite{google2025gemini3flashpreview}) & 69.4 & 56.2 & 20.8 & 38.9 & 46.3 & 63.8 & 44.9 & 20.8 & 11.7 & 35.3 \\
Gemini-2.5-Pro (\cite{comanici2025gemini25}) & 87.2 & 59.2 & 40.4 & 46.8 & 58.4 & 24.5 & 43.0 & 45.3 & 51.7 & 41.1 \\
Qwen-3.5-Plus (\cite{qwen2026qwen35})& 84.5 & 55.1 & 44.5 & 58.9 & 60.8 & 53.2 & 56.6 & 44.5 & 39.6 & 48.5 \\
Qwen3-VL-Flash (\cite{bai2025qwen3vl})& 65.5 & 49.5 & 39.0 & 55.8 & 52.5 & 93.8 & 50.0 & 27.8 & 47.2 & 54.7 \\
\midrule
\multicolumn{11}{l}{\textit{Open-Source Models}} \\
Intern-S1 (\cite{bai2025interns1}) & 83.8 & 63.8 & 42.3 & 44.2 & 58.5 & 82.3 & 61.5 & 60.8 & 34.3 & 59.7 \\
Intern-S1-Mini (\cite{bai2025interns1})& 47.2 & 52.1 & 29.8 & 43.4 & 43.1 & 55.8 & 55.1 & 41.9 & 41.1 & 48.5 \\
Qwen-3-VL-30B-A3B-Instruct (\cite{bai2025qwen3vl})& 72.1 & 44.2 & 27.5 & 43.4 & 46.8 & 63.8 & 45.7 & 40.8 & 53.2 & 50.9 \\
Qwen-2.5-VL-72B-Instruct (\cite{bai2025qwen25vl})& 72.1 & 65.3 & 51.3 & 41.9 & 57.7 & 71.7 & 58.1 & 63.0 & 46.0 & 59.7 \\
V2X-MoE-8B (Ours) & \textbf{98.9} & \textbf{94.3} & \textbf{93.2} & \textbf{94.7} & \textbf{95.3} & \textbf{99.6} & \textbf{95.1} & \textbf{92.1} & \textbf{89.1} & \textbf{94.0} \\
\bottomrule
\end{tabular}
\end{table*}

\subsubsection{Cooperative results}
\label{sec:results_co}

Table~\ref{tab:main_co_results} reports the benchmark results on the four cooperative tasks. Compared with the single-view settings above, cooperative evaluation reveals a much more challenging and unstable performance profile for general-purpose MLLMs. The best average among the proprietary models is only 36.7, achieved by both GPT-5.2 and Gemini-3-Flash-Preview, while the strongest open-source baseline, Intern-S1, reaches 43.5. These values are markedly lower than the corresponding vehicle-side and infrastructure-side averages, showing that cooperative reasoning is not obtained automatically by simply providing two views.

The task-level breakdown further illustrates this difficulty. Many models exhibit highly uneven behavior across the four cooperative tasks. For example, Gemini-3-Flash-Preview performs very strongly on CO1 but collapses on CO2 and CO4, Gemini-2.5-Pro reaches its best score on CO2 but remains extremely weak on CO1 and CO4, and Qwen-3.5-Plus attains its strongest result on CO3 while again dropping sharply on CO4. Similar instability is also visible among open-source models. These patterns indicate that cooperative reasoning in V2X-QA is not a single capability, but a compound challenge involving cross-view alignment, visibility reasoning, motion anticipation, and planning-oriented decision making.

Against this backdrop, V2X-MoE again achieves the strongest overall result by a large margin. Its cooperative average reaches 86.2, with consistently strong performance across all four cooperative tasks, including 92.1 on CO1, 93.2 on CO2, 83.4 on CO3, and 75.9 on CO4. In contrast to the pronounced task-specific collapses observed in the off-the-shelf models, V2X-MoE remains much more balanced across the cooperative benchmark. This result provides the strongest evidence for the effectiveness of the proposed design, because cooperative reasoning is precisely the regime in which viewpoint-conditioned routing and specialization are most needed.

At a higher level, the cooperative results reinforce two conclusions. First, the cooperative setting of V2X-QA is substantially more demanding than either vehicle-side or infrastructure-side reasoning for current general-purpose MLLMs. Second, explicit viewpoint-aware adaptation matters: the large gap between V2X-MoE and the non-specialized baselines suggests that successful cooperative reasoning requires more than raw model capacity or an additional image input. Instead, it benefits from architectural bias and training strategy that are explicitly aligned with cross-view evidence integration.

\begin{table*}[t]
\centering
\caption{Main benchmark results on the cooperative tasks of V2X-QA. Accuracy (\%) is reported for each task and the average over the cooperative viewpoint group.}
\label{tab:main_co_results}
\setlength{\tabcolsep}{7pt}
\renewcommand{\arraystretch}{1.15}
\small
\begin{tabular}{lccccc}
\toprule
Model & CO1 & CO2 & CO3 & CO4 & Avg. \\
\midrule
\multicolumn{6}{l}{\textit{Closed-Source Models}} \\
GPT-5.2 (\cite{openai2025gpt52}) & 34.0 & 36.2 & 49.4 & 27.2 & 36.7 \\
GPT-5-Mini (\cite{openai2025gpt5mini})& 26.4 & 9.8 & 47.2 & 11.3 & 23.7 \\
Gemini-3-Flash-Preview (\cite{google2025gemini3flashpreview}) & 86.0 & 3.8 & 43.4 & 13.6 & 36.7 \\
Gemini-2.5-Pro (\cite{comanici2025gemini25}) & 15.8 & 59.2 & 45.3 & 3.4 & 30.9 \\
Qwen-3.5-Plus (\cite{qwen2026qwen35}) & 37.7 & 30.9 & 64.5 & 2.8 & 34.0 \\
Qwen3-VL-Flash (\cite{bai2025qwen3vl}) & 22.3 & 47.9 & 44.9 & 6.5 & 30.4 \\
\midrule
\multicolumn{6}{l}{\textit{Open-Source Models}} \\
Intern-S1 (\cite{bai2025interns1}) & 57.4 & 70.2 & 43.0 & 3.4 & 43.5 \\
Intern-S1-Mini (\cite{bai2025interns1}) & 64.5 & 29.1 & 43.4 & 5.3 & 35.6 \\
Qwen-3-VL-30B-A3B-Instruct (\cite{bai2025qwen3vl}) & 20.8 & 32.1 & 52.1 & 4.9 & 27.5 \\
Qwen-2.5-VL-72B-Instruct (\cite{bai2025qwen25vl}) & 61.9 & 54.7 & 42.6 & 8.7 & 42.0 \\
V2X-MoE-8B (Ours) & \textbf{92.1} & \textbf{93.2} & \textbf{83.4} & \textbf{75.9} & \textbf{86.2} \\
\bottomrule
\end{tabular}
\end{table*}

Taken together, Tables~\ref{tab:main_vs_is_results} and~\ref{tab:main_co_results} support three main conclusions. First, off-the-shelf proprietary and open-source MLLMs remain far from saturated on V2X-QA, especially under the cooperative setting. Second, viewpoint matters: vehicle-side, infrastructure-side, and cooperative settings expose different capability bottlenecks rather than forming a simple monotonic difficulty ordering. Third, the proposed V2X-MoE baseline demonstrates that strong benchmark performance can be obtained when the model is explicitly adapted to viewpoint-conditioned MCQA reasoning. In particular, its large margin over the compared baselines on all three viewpoint groups validates the usefulness of explicit view routing, viewpoint-specific expert adaptation, and staged refinement under the V2X-QA benchmark.

\subsection{Diagnostic and reliability analysis}
\label{sec:diagnostic_reliability}

Beyond the main benchmark tables, V2X-QA is designed to reveal structured differences in model behavior across viewpoints and task families. In this subsection, we first analyze task-level performance patterns and then report supplementary reliability analysis for the V2X-MoE baseline.

\subsubsection{Diagnostic analysis across views and tasks}
\label{sec:diagnostic_views_tasks}

\begin{figure}[pos=htbp]
    \centering
    \includegraphics[width=\textwidth]{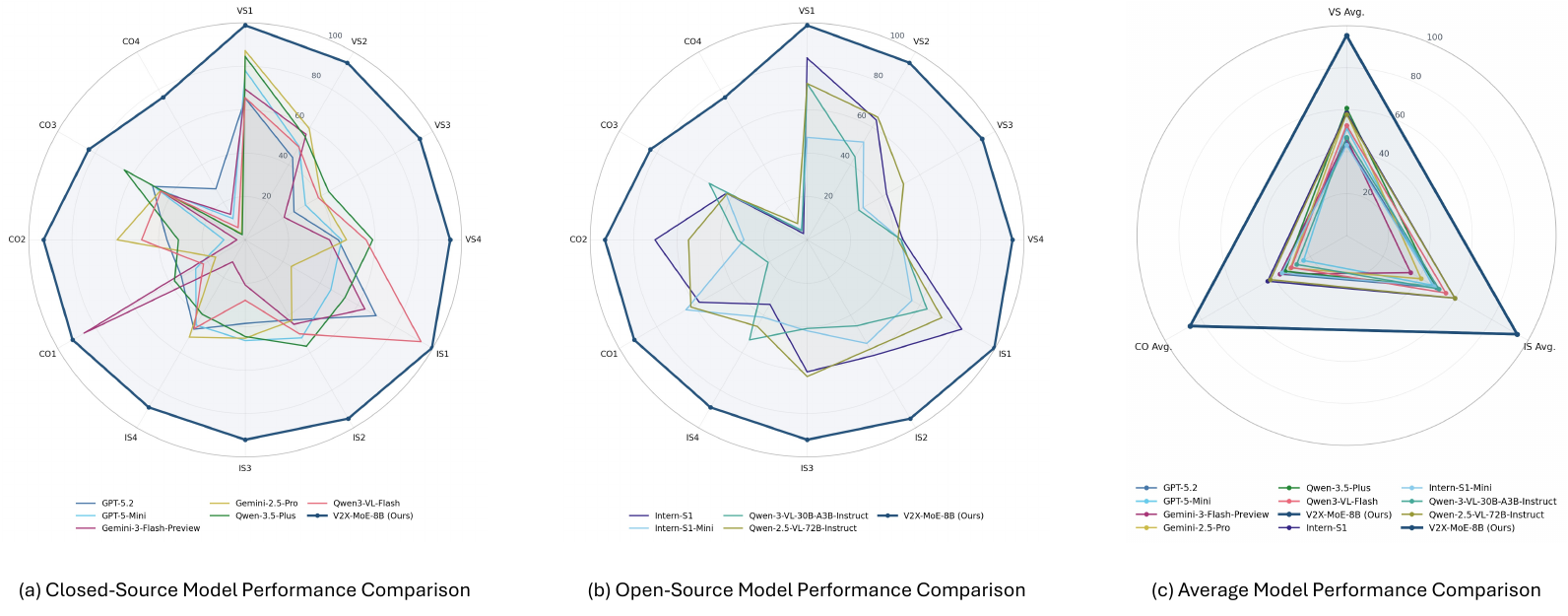}
    \caption{Task-level and viewpoint-level model performance comparison on V2X-QA.}
    \label{fig:radar_results}
\end{figure}

Figure~\ref{fig:radar_results} visualizes task-level performance patterns for representative proprietary and open-source models, together with viewpoint-group averages. Compared with the aggregated results in Section~\ref{sec:main_results}, the radar plots make the diagnostic role of V2X-QA more explicit by showing not only which model performs better overall, but also how performance is distributed across the twelve tasks and the three viewpoint groups.

Several patterns are immediately visible. First, the task difficulty structure varies substantially across viewpoints rather than shifting uniformly. Vehicle-side tasks emphasize local scene semantics, visible agents, ego-relevant risk, and short-horizon prediction from onboard observations. Infrastructure-side tasks instead emphasize global layout, broader traffic organization, macroscopic risk, and long-range cues available from the roadside perspective. Cooperative tasks impose an additional burden: the model must not only recognize information from each view, but also integrate and reconcile complementary evidence across views. As a result, cooperative tasks exhibit the strongest instability for general-purpose MLLMs, especially on planning-oriented CO4 and, for some models, also on visibility- and occlusion-related CO2.

Second, the radar plots reveal pronounced differences in model profile. Many proprietary and open-source MLLMs show highly uneven task-wise behavior, with sharp peaks on some tasks but severe collapses on others. This unevenness is especially clear in the cooperative setting, where several strong off-the-shelf models still fail to maintain stable performance across CO1--CO4. By contrast, V2X-MoE exhibits a much more regular and consistently strong profile across the full twelve-task space. Its gains are not restricted to a single subset, but extend across vehicle-side, infrastructure-side, and cooperative tasks, which is consistent with the intended effect of explicit view routing, view-specialized LoRA adaptation, and staged refinement.

Third, the average-view radar plot reinforces the same conclusion at a more compact level. The compared proprietary and open-source models form a relatively clustered band, with limited separation between their VS, IS, and CO averages. V2X-MoE stands clearly outside this band and remains strong under all three evidence conditions, indicating that the proposed benchmark-aligned specialization improves not only a few isolated tasks but also the overall viewpoint-conditioned reasoning profile of the model.

\subsubsection{Reliability analysis for V2X-MoE}
\label{sec:baseline_reliability}

We next report reliability analysis for V2X-MoE as a supplementary diagnostic. This analysis is restricted to the baseline rather than applied uniformly to all benchmarked models, because it relies on a controlled probabilistic interface derived from the V2X-MoE inference pipeline introduced in Section~\ref{sec:exp_setup}. The purpose here is not to redefine the benchmark around calibration metrics, but to examine whether confidence quality changes systematically across viewpoint conditions even when the model architecture and evaluation protocol are fixed.

Table~\ref{tab:calibration_results} summarizes ECE and Brier score for the vehicle-side, infrastructure-side, and cooperative subsets. Two patterns are immediately apparent. First, the vehicle-side and infrastructure-side subsets exhibit very similar calibration quality, with ECE values of 0.0386 and 0.0387, respectively. Their Brier scores are also relatively low, indicating that under single-view reasoning the model's confidence remains broadly aligned with empirical correctness. Second, the cooperative subset is noticeably less reliable, with the highest ECE of 0.0865 and a substantially larger Brier score of 0.2292. This result is consistent with the main benchmark findings: cooperative reasoning is not only more difficult in terms of accuracy, but also more challenging from the perspective of uncertainty estimation.

\begin{figure}[pos=htbp]
    \centering
    \includegraphics[width=0.75\columnwidth]{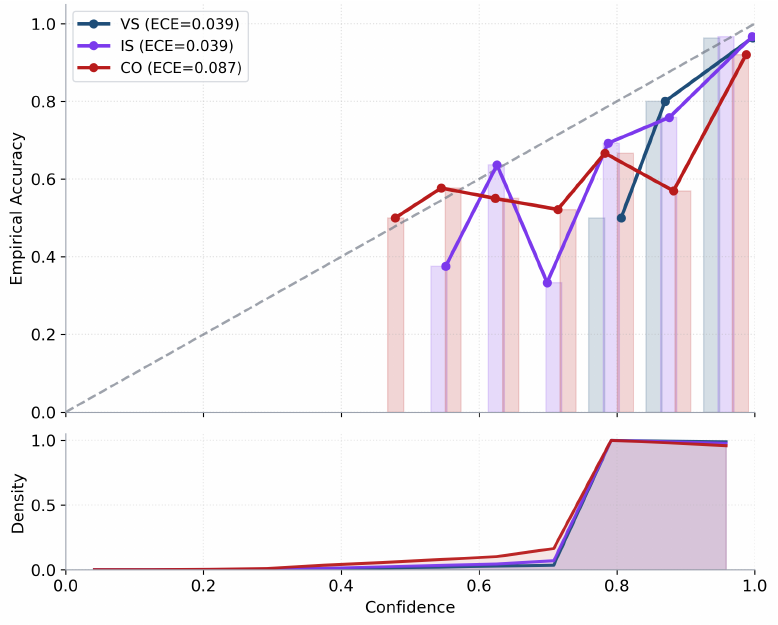}
    \caption{Reliability diagram and confidence distribution of V2X-MoE across vehicle-side, infrastructure-side, and cooperative viewpoint groups.}
    \label{fig:reliability_results}
\end{figure}

These trends are further illustrated in Figure~\ref{fig:reliability_results}, which plots the reliability diagram together with the confidence distribution. The VS and IS curves remain comparatively close to the diagonal across the occupied confidence range, whereas the CO curve departs more visibly from ideal calibration. The lower panel further shows that confidence mass is concentrated in the high-confidence region across all three viewpoint groups, but the cooperative setting still exhibits a larger mismatch between confidence and empirical accuracy. This indicates that the main challenge in the cooperative regime is not merely making the correct selection, but doing so with confidence that faithfully reflects uncertainty.

\begin{table}[t]
\centering
\caption{Calibration results of V2X-MoE across viewpoint groups.}
\label{tab:calibration_results}
\setlength{\tabcolsep}{8pt}
\renewcommand{\arraystretch}{1.15}
\small
\begin{tabular}{lccc}
\toprule
View & ECE & Brier & No. of Samples \\
\midrule
VS  & 0.0386 & 0.0854 & 1{,}060 \\
IS  & 0.0387 & 0.0963 & 1{,}060 \\
CO  & 0.0865 & 0.2292 & 1{,}060 \\
All & 0.0530 & 0.1370 & 3{,}180 \\
\bottomrule
\end{tabular}
\end{table}

\section{Discussion}
\label{sec:discussion}

Experiment results show that V2X-QA captures a substantially broader reasoning problem than conventional driving QA benchmarks. Vehicle-side, infrastructure-side, and cooperative evaluation are not minor variants of the same task, but distinct observability regimes with different reasoning requirements. In particular, the infrastructure-side setting forms an independent regime for macroscopic traffic interpretation, global layout understanding, and long-range cue utilization, while the cooperative setting is more demanding because it requires the model to reconcile complementary evidence across views and convert it into coherent prediction- and planning-oriented reasoning. The strong performance of V2X-MoE further suggests that explicit viewpoint specialization is an effective inductive bias for this benchmark. Taken together, these findings support the central motivation of V2X-QA: conclusions about cooperative autonomous driving reasoning should not be drawn from ego-centric evaluation alone, and viewpoint accessibility should be treated as an explicit experimental variable rather than an implicit assumption.

At the same time, several limitations should be noted. Although V2X-QA provides a real-world, expert-verified, and view-decoupled benchmark, it remains an image-based MCQA benchmark rather than a closed-loop driving evaluation framework. The cooperative setting is currently defined over paired vehicle-side and infrastructure-side observations, and therefore does not yet capture richer communication effects such as latency, packet loss, asynchronous sensing, or dynamically changing sensor availability. In addition, the calibration analysis in this work is limited to V2X-MoE and is intended as a supplementary diagnosis rather than a benchmark-wide comparison metric. Future work can extend V2X-QA toward video-based temporal reasoning, more diverse cities and traffic scenarios, communication-aware cooperative settings, and broader reliability analysis across model families.

\section{Conclusion}
\label{sec:conclusion}

This work presents V2X-QA, a novel benchmark for studying multimodal reasoning in autonomous driving under vehicle-side, infrastructure-side, and cooperative viewpoints. Unlike conventional driving QA settings that mainly focus on ego-view evidence, V2X-QA explicitly treats viewpoint accessibility as part of the problem formulation and enables controlled comparison across different evidence conditions within a unified MCQA framework. Through its twelve-task design, the benchmark provides a structured way to examine how current MLLMs handle scene understanding, prediction, and reasoning and planning under heterogeneous observability.

Our experiments show that viewpoint changes are not a superficial variation of the same task, but a meaningful source of capability difference. In particular, the infrastructure-side setting captures important macroscopic traffic understanding that is difficult to recover from ego observations alone, while the cooperative setting exposes the unresolved challenge of cross-view alignment and evidence integration. The strong performance of V2X-MoE indicates that explicitly modeling viewpoint structure is an effective direction for this problem. We envision V2X-QA can serve as a useful benchmark for future research on multi-perspective reasoning, cooperative embodied intelligence, and more reliable multimodal systems for connected intelligent driving.

\appendix
\section*{Appendix}
\addcontentsline{toc}{section}{Appendix}

\section{MCQA task bank used for annotation}
\label{app:mcqa_task_bank}

This appendix presents the complete MCQA task bank used during annotation. The task bank is organized into three parts corresponding to VS, IS, and CO tasks. Tables~\ref{tab:appendix_vs_task_bank}, \ref{tab:appendix_is_task_bank}, and \ref{tab:appendix_co_task_bank} provide the full task banks for the VS, IS, and CO groups, respectively. Each entry includes the question ID, question text, candidate options, and the corresponding canonical answers.

\newcolumntype{M}[1]{>{\raggedright\arraybackslash}m{#1}}
\newcolumntype{s}{>{\hsize=0.75\hsize\raggedright\arraybackslash}X}
\newcolumntype{b}{>{\hsize=1.25\hsize\raggedright\arraybackslash}X}

\renewcommand{\arraystretch}{1.08}
\setlength{\tabcolsep}{2pt}
\setlength\LTleft{0pt}
\setlength\LTright{0pt}
\scriptsize

\begin{xltabular}{\textwidth}{@{}M{0.100\textwidth}M{0.080\textwidth}M{0.12\textwidth}s b@{}}
\caption{Vehicle-side MCQA task bank for VS tasks.}
\label{tab:appendix_vs_task_bank}\\
\toprule
Task & Question ID & Question & Options & Canonical Answers \\
\midrule
\endfirsthead

\toprule
Task & Question ID & Question & Options & Canonical Answers \\
\midrule
\endhead

\bottomrule
\endfoot

\multirow[c]{3}{0.115\textwidth}{VS1:\newline Ego Static Scene}
& VS1\_Q1
& What is the main road scene ahead?
& A. An intersection area.\newline
B. A straight road segment.\newline
C. A work-zone or barrier area.\newline
D. A merge or ramp area.
& A: The main road scene ahead is an intersection area.\newline
B: The main road scene ahead is a straight road segment.\newline
C: The main road scene ahead is a work-zone or barrier area.\newline
D: The main road scene ahead is a merge or ramp area. \\

& VS1\_Q2
& Which road feature is most important ahead?
& A. Junction structure.\newline
B. Lane-following road.\newline
C. Barrier or cone guidance.\newline
D. Ramp-like road split.
& A: The most important road feature ahead is a junction structure.\newline
B: The most important road feature ahead is a lane-following road.\newline
C: The most important road feature ahead is barrier or cone guidance.\newline
D: The most important road feature ahead is a ramp-like road split. \\

& VS1\_Q3
& How should the scene ahead be grouped?
& A. Intersection-type scene.\newline
B. Regular road scene.\newline
C. Restricted or guided scene.\newline
D. Branching road scene.
& A: The scene ahead should be grouped as an intersection-type scene.\newline
B: The scene ahead should be grouped as a regular road scene.\newline
C: The scene ahead should be grouped as a restricted or guided scene.\newline
D: The scene ahead should be grouped as a branching road scene. \\
\midrule

\multirow[c]{3}{0.115\textwidth}{VS2:\newline Ego Visible Agents}
& VS2\_Q1
& What is the main road user ahead?
& A. No clear road user ahead.\newline
B. A car-like vehicle.\newline
C. A large vehicle.\newline
D. A person, bike, or motorcycle.
& A: There is no clear road user ahead.\newline
B: The main road user ahead is a car-like vehicle.\newline
C: The main road user ahead is a large vehicle.\newline
D: The main road user ahead is a person, bike, or motorcycle. \\

& VS2\_Q2
& Which category best fits the key target ahead?
& A. None is clearly key.\newline
B. Passenger vehicle.\newline
C. Bus or truck.\newline
D. Small or vulnerable user.
& A: No target is clearly key ahead.\newline
B: The key target ahead is a passenger vehicle.\newline
C: The key target ahead is a bus or truck.\newline
D: The key target ahead is a small or vulnerable road user. \\

& VS2\_Q3
& Who matters most for the ego path ahead?
& A. No one clearly matters most.\newline
B. A car or SUV.\newline
C. A truck or bus.\newline
D. A rider, cyclist, or pedestrian.
& A: No road user clearly matters most for the ego path ahead.\newline
B: A car or SUV matters most for the ego path ahead.\newline
C: A truck or bus matters most for the ego path ahead.\newline
D: A rider, cyclist, or pedestrian matters most for the ego path ahead. \\
\midrule

\multirow[c]{3}{0.115\textwidth}{VS3:\newline Ego Risk}
& VS3\_Q1
& What is the main risk ahead?
& A. Cross traffic.\newline
B. A leading vehicle.\newline
C. A vulnerable road user.\newline
D. No clear risk.
& A: The main risk ahead is cross traffic.\newline
B: The main risk ahead is a leading vehicle.\newline
C: The main risk ahead is a vulnerable road user.\newline
D: There is no clear risk ahead. \\

& VS3\_Q2
& Which hazard is most important now?
& A. Blurred visibility or wet road surface.\newline
B. Vehicle conflict.\newline
C. Pedestrian or rider conflict.\newline
D. No obvious hazard.
& A: The most important hazard now is blurred visibility or a wet road surface.\newline
B: The most important hazard now is a front or side vehicle conflict.\newline
C: The most important hazard now is a pedestrian or rider conflict.\newline
D: There is no obvious hazard now. \\

& VS3\_Q3
& What should the ego pay most attention to?
& A. Traffic from the side.\newline
B. Traffic directly ahead.\newline
C. People or two-wheelers nearby.\newline
D. Nothing stands out strongly.
& A: The ego should pay most attention to traffic from the side.\newline
B: The ego should pay most attention to traffic directly ahead.\newline
C: The ego should pay most attention to people or two-wheelers nearby.\newline
D: Nothing stands out strongly for the ego to attend to. \\
\midrule

\multirow[c]{3}{0.115\textwidth}{VS4:\newline Ego Prediction}
& VS4\_Q1
& What will the ego vehicle likely do next?
& A. Go straight.\newline
B. Slow down.\newline
C. Turn or change lanes.\newline
D. It is unclear.
& A: The ego vehicle will likely go straight next.\newline
B: The ego vehicle will likely slow down next.\newline
C: The ego vehicle will likely turn or change lanes next.\newline
D: The next action of the ego vehicle is unclear. \\

& VS4\_Q2
& How is the ego vehicle likely to move?
& A. Keep moving forward.\newline
B. Make a left turn.\newline
C. Make a right turn.\newline
D. Change lanes.
& A: The ego vehicle is likely to keep moving forward.\newline
B: The ego vehicle is likely to make a left turn.\newline
C: The ego vehicle is likely to make a right turn.\newline
D: The ego vehicle is likely to change lanes. \\

& VS4\_Q3
& What is the most likely next action of the ego vehicle?
& A. Go straight.\newline
B. Turn left.\newline
C. Turn right.\newline
D. Change lanes.
& A: The most likely action is to go straight.\newline
B: The most likely action is to turn left.\newline
C: The most likely action is to turn right.\newline
D: The most likely action is to change lanes. \\

\end{xltabular}
\normalsize

\newcolumntype{M}[1]{>{\raggedright\arraybackslash}m{#1}}
\newcolumntype{s}{>{\hsize=0.85\hsize\raggedright\arraybackslash}X}
\newcolumntype{b}{>{\hsize=1.15\hsize\raggedright\arraybackslash}X}

\renewcommand{\arraystretch}{1.08}
\setlength{\tabcolsep}{2pt}
\setlength\LTleft{0pt}
\setlength\LTright{0pt}
\scriptsize

\begin{xltabular}{\textwidth}{@{}M{0.100\textwidth}M{0.080\textwidth}M{0.12\textwidth}s b@{}}
\caption{Infrastructure-side MCQA task bank for IS tasks.}
\label{tab:appendix_is_task_bank}\\
\toprule
Task & Question ID & Question & Options & Canonical Answers \\
\midrule
\endfirsthead

\toprule
Task & Question ID & Question & Options & Canonical Answers \\
\midrule
\endhead

\bottomrule
\endfoot

\multirow[c]{3}{0.115\textwidth}{IS1:\newline RSU Global Layout}
& IS1\_Q1
& What is the most prominent layout feature in this RSU view?
& A. A regular intersection.\newline
B. A curved road approach.\newline
C. Strong turn guidance.\newline
D. A complex layout.
& A: The most prominent layout feature in this RSU view is a regular intersection.\newline
B: The most prominent layout feature in this RSU view is a curved road approach.\newline
C: The most prominent layout feature in this RSU view is strong turn guidance.\newline
D: The most prominent layout feature in this RSU view is a complex layout. \\

& IS1\_Q2
& Which intersection feature stands out most in this RSU view?
& A. A regular intersection shape.\newline
B. A curved approach road.\newline
C. Clear turn guidance.\newline
D. A complex road layout.
& A: The most prominent feature in this RSU view is a regular intersection shape.\newline
B: The most prominent feature in this RSU view is a curved approach road.\newline
C: The most prominent feature in this RSU view is clear turn guidance.\newline
D: The most prominent feature in this RSU view is a complex road layout. \\

& IS1\_Q3
& How should this RSU-view intersection be described?
& A. A standard intersection.\newline
B. An intersection with a curved approach.\newline
C. An intersection with strong turn guidance.\newline
D. An intersection with a complex layout.
& A: This RSU-view intersection is best described as a standard intersection.\newline
B: This RSU-view intersection is best described as an intersection with a curved approach.\newline
C: This RSU-view intersection is best described as an intersection with strong turn guidance.\newline
D: This RSU-view intersection is best described as an intersection with a complex layout. \\
\midrule

\multirow[c]{3}{0.115\textwidth}{IS2:\newline RSU Traffic Agents}
& IS2\_Q1
& What best describes the traffic in the RSU view?
& A. Light traffic.\newline
B. Many cars close together.\newline
C. Mixed traffic with vulnerable users.\newline
D. A large vehicle stands out.
& A: The RSU view shows light traffic.\newline
B: The RSU view shows many cars close together.\newline
C: The RSU view shows mixed traffic with vulnerable users.\newline
D: The RSU view is dominated by a large vehicle. \\

& IS2\_Q2
& Which traffic pattern is most clear?
& A. Sparse movement.\newline
B. Dense vehicle group.\newline
C. Mixed road users.\newline
D. One large vehicle.
& A: The clearest traffic pattern is sparse movement.\newline
B: The clearest traffic pattern is a dense vehicle group.\newline
C: The clearest traffic pattern is mixed road users.\newline
D: The clearest traffic pattern is one large vehicle. \\

& IS2\_Q3
& What is the key traffic feature here?
& A. Few active agents.\newline
B. Clustered vehicles.\newline
C. Different user types.\newline
D. A standout truck or bus.
& A: The key traffic feature is few active agents.\newline
B: The key traffic feature is clustered vehicles.\newline
C: The key traffic feature is different user types.\newline
D: The key traffic feature is a standout truck or bus. \\
\midrule

\multirow[c]{3}{0.115\textwidth}{IS3:\newline RSU Global Risk}
& IS3\_Q1
& What is the main global risk?
& A. A wet road surface.\newline
B. Dense vehicle interaction.\newline
C. Vulnerable road users.\newline
D. No clear global risk.
& A: The main global risk is a wet road surface.\newline
B: The main global risk is dense vehicle interaction.\newline
C: The main global risk is vulnerable road users.\newline
D: There is no clear global risk. \\

& IS3\_Q2
& Which risk matters most in the RSU view?
& A. A wet slippery surface.\newline
B. Crowded traffic flow.\newline
C. Pedestrian or rider exposure.\newline
D. Nothing stands out strongly.
& A: The most important RSU-view risk is a wet slippery surface.\newline
B: The most important RSU-view risk is crowded traffic flow.\newline
C: The most important RSU-view risk is pedestrian or rider exposure.\newline
D: No risk stands out strongly in the RSU view. \\

& IS3\_Q3
& What deserves the most caution here?
& A. A wet road surface.\newline
B. Heavy interaction among vehicles.\newline
C. Nearby vulnerable users.\newline
D. No major caution point.
& A: The main caution point is a wet road surface.\newline
B: The main caution point is heavy interaction among vehicles.\newline
C: The main caution point is nearby vulnerable users.\newline
D: There is no major caution point. \\
\midrule

\multirow[c]{3}{0.115\textwidth}{IS4:\newline RSU Long-Range Cues}
& IS4\_Q1
& What is the clearest long-range cue?
& A. A far large vehicle.\newline
B. A far vehicle group.\newline
C. A work zone ahead.\newline
D. No strong long-range cue.
& A: The clearest long-range cue is a far large vehicle.\newline
B: The clearest long-range cue is a far vehicle group.\newline
C: The clearest long-range cue is a work zone ahead.\newline
D: There is no strong long-range cue. \\

& IS4\_Q2
& Which distant feature stands out most?
& A. A distant truck or bus.\newline
B. A distant traffic cluster.\newline
C. A distant work zone.\newline
D. Nothing distant stands out.
& A: The most salient distant feature is a truck or bus.\newline
B: The most salient distant feature is a traffic cluster.\newline
C: The most salient distant feature is a work zone.\newline
D: No distant feature stands out. \\

& IS4\_Q3
& What useful far-field cue is visible?
& A. A large far-field agent.\newline
B. A far buildup of traffic.\newline
C. A far work zone.\newline
D. No obvious far-field cue.
& A: The useful far-field cue is a large far-field agent.\newline
B: The useful far-field cue is a far buildup of traffic.\newline
C: The useful far-field cue is a far work zone.\newline
D: There is no obvious far-field cue. \\

\end{xltabular}
\normalsize

\newcolumntype{M}[1]{>{\raggedright\arraybackslash}m{#1}}
\newcolumntype{s}{>{\hsize=0.82\hsize\raggedright\arraybackslash}X}
\newcolumntype{b}{>{\hsize=1.18\hsize\raggedright\arraybackslash}X}

\renewcommand{\arraystretch}{1.08}
\setlength{\tabcolsep}{2pt}
\setlength\LTleft{0pt}
\setlength\LTright{0pt}
\scriptsize

\begin{xltabular}{\textwidth}{@{}M{0.100\textwidth}M{0.080\textwidth}M{0.12\textwidth}s b@{}}
\caption{Cooperative MCQA task bank for CO tasks.}
\label{tab:appendix_co_task_bank}\\
\toprule
Task & Question ID & Question & Options & Canonical Answers \\
\midrule
\endfirsthead

\toprule
Task & Question ID & Question & Options & Canonical Answers \\
\midrule
\endhead

\bottomrule
\endfoot

\multirow[c]{3}{0.115\textwidth}{CO1:\newline Cooperative \newline Scene Understanding}
& CO1\_Q1
& With both views, what best describes the ego path?
& A. The path looks clear.\newline
B. Cross traffic affects the path.\newline
C. The path is constrained.\newline
D. The path is still unclear.
& A: With both views, the ego path looks clear.\newline
B: With both views, cross traffic affects the ego path.\newline
C: With both views, the ego path is constrained.\newline
D: With both views, the ego path is still unclear. \\

& CO1\_Q2
& What is the main joint scene result?
& A. Open path ahead.\newline
B. Crossing activity matters.\newline
C. The lane area is limited.\newline
D. Joint evidence remains weak.
& A: The main joint scene result is an open path ahead.\newline
B: The main joint scene result is that crossing activity matters.\newline
C: The main joint scene result is that the lane area is limited.\newline
D: The main joint scene result is that the joint evidence remains weak. \\

& CO1\_Q3
& How should the cooperative scene be summarized?
& A. Mostly clear forward scene.\newline
B. Cross-view conflict scene.\newline
C. Restricted-path scene.\newline
D. Uncertain combined scene.
& A: The cooperative scene is best summarized as a mostly clear forward scene.\newline
B: The cooperative scene is best summarized as a cross-view conflict scene.\newline
C: The cooperative scene is best summarized as a restricted-path scene.\newline
D: The cooperative scene is best summarized as an uncertain combined scene. \\
\midrule

\multirow[c]{3}{0.115\textwidth}{CO2:\newline Cooperative Visibility}
& CO2\_Q1
& What does cooperation add most?
& A. An occluded road user becomes clear.\newline
B. A blurred ego-view scene becomes clear.\newline
C. A long-range cue becomes clear.\newline
D. It adds little new information.
& A: Cooperation mainly makes an occluded road user clear.\newline
B: Cooperation mainly makes a blurred ego-view scene clear.\newline
C: Cooperation mainly makes a long-range cue clear.\newline
D: Cooperation adds little new information. \\

& CO2\_Q2
& Which missing cue is best recovered by cooperation?
& A. An occluded road-user cue.\newline
B. A cue missing from the blurred ego view.\newline
C. A long-range cue.\newline
D. No important cue is recovered.
& A: The best recovered missing cue is an occluded road-user cue.\newline
B: The best recovered missing cue is a cue missing from the blurred ego view.\newline
C: The best recovered missing cue is a long-range cue.\newline
D: No important cue is recovered. \\

& CO2\_Q3
& What is the main value of the second view?
& A. It reveals an occluded road user.\newline
B. It clarifies a blurred ego view.\newline
C. It provides a long-range cue.\newline
D. Its added value is limited.
& A: The main value of the second view is that it reveals an occluded road user.\newline
B: The main value of the second view is that it clarifies a blurred ego view.\newline
C: The main value of the second view is that it provides a long-range cue.\newline
D: The added value of the second view is limited. \\
\midrule

\multirow[c]{3}{0.115\textwidth}{CO3:\newline Cooperative Prediction}
& CO3\_Q1
& With both views, what will the ego vehicle do next?
& A. Go straight.\newline
B. Make a left turn.\newline
C. Make a right turn.\newline
D. Change lanes.
& A: With both views, the ego vehicle will likely go straight next.\newline
B: With both views, the ego vehicle will likely make a left turn next.\newline
C: With both views, the ego vehicle will likely make a right turn next.\newline
D: With both views, the ego vehicle will likely change lanes. \\

& CO3\_Q2
& What motion is most likely for the ego vehicle after combining views?
& A. Go straight.\newline
B. Make a left turn.\newline
C. Make a right turn.\newline
D. Change lanes.
& A: After combining views, the ego vehicle is most likely to go straight.\newline
B: After combining views, the ego vehicle is most likely to make a left turn.\newline
C: After combining views, the ego vehicle is most likely to make a right turn.\newline
D: After combining views, the ego vehicle is most likely to change lanes. \\

& CO3\_Q3
& What is the best cooperative prediction for the ego vehicle?
& A. Go straight.\newline
B. Make a left turn.\newline
C. Make a right turn.\newline
D. Change lanes.
& A: The best short-term cooperative prediction for the ego vehicle is to go straight.\newline
B: The best short-term cooperative prediction for the ego vehicle is to make a left turn.\newline
C: The best short-term cooperative prediction for the ego vehicle is to make a right turn.\newline
D: The best short-term cooperative prediction for the ego vehicle is to change lanes. \\
\midrule

\multirow[c]{3}{0.115\textwidth}{CO4:\newline Cooperative Planning}
& CO4\_Q1
& With both views, what should the ego vehicle do immediately?
& A. Accelerate.\newline
B. Reduce speed.\newline
C. Keep the current speed.\newline
D. Yield or prepare to stop.
& A: With both views, the ego vehicle should accelerate.\newline
B: With both views, the ego vehicle should reduce speed.\newline
C: With both views, the ego vehicle should keep the current speed.\newline
D: With both views, the ego vehicle should yield or prepare to stop. \\

& CO4\_Q2
& With both views, what is the best immediate ego action?
& A. Accelerate.\newline
B. Reduce speed.\newline
C. Keep the current speed.\newline
D. Yield or prepare to stop.
& A: With both views, the best immediate ego action is to accelerate.\newline
B: With both views, the best immediate ego action is to reduce speed.\newline
C: With both views, the best immediate ego action is to keep the current speed.\newline
D: With both views, the best immediate ego action is to yield or prepare to stop. \\

& CO4\_Q3
& With both views, what is the best immediate planning choice?
& A. Pick up speed.\newline
B. Slow down.\newline
C. Maintain the current speed.\newline
D. Yield or get ready to stop.
& A: With both views, the best immediate planning choice is to pick up speed.\newline
B: With both views, the best immediate planning choice is to slow down.\newline
C: With both views, the best immediate planning choice is to maintain the current speed.\newline
D: With both views, the best immediate planning choice is to yield or get ready to stop. \\

\end{xltabular}
\normalsize

\section{Unified benchmark prompt template for standardized MCQA evaluation}
\label{app:benchmark_prompt}

As described in the main text, all benchmark comparisons in V2X-QA are conducted under a standardized and unified MCQA evaluation protocol to ensure fair, reproducible, and directly comparable benchmarking across different models and viewpoint conditions. To support this protocol, we design a unified benchmark prompt template that is used consistently for all evaluated models, including proprietary MLLMs, open-source MLLMs, and V2X-MoE. Table~\ref{tab:appendix_unified_benchmark_prompt} summarizes the complete prompt structure, including the fixed system prompt, the shared user prompt template, and the viewpoint-specific image notes for the VS, IS, and CO settings.

\renewcommand{\arraystretch}{1.12}
\setlength{\tabcolsep}{4pt}
\scriptsize

\begin{table}[htbp]
\centering
\caption{Unified benchmark prompt template used for standardized MCQA evaluation across all benchmarked models.}
\label{tab:appendix_unified_benchmark_prompt}
\begin{tabular}{m{0.23\textwidth} m{0.70\textwidth}}
\toprule
Component & Content \\
\midrule

System prompt
&
You are answering a multiple-choice autonomous driving question.\newline
Use only the provided image evidence and the question/options.\newline
Return exactly one uppercase letter only: A or B or C or D. Do not output any other words, punctuation, or explanation.
\\
\midrule

User prompt template
&
Task: \{task\_id\}\newline
Viewpoint: \{viewpoint\}\newline

Image evidence:\newline
\{image\_note\}\newline

Question:\newline
\{question\}\newline

Options:\newline
A. \{opt\_a\}\newline
B. \{opt\_b\}\newline
C. \{opt\_c\}\newline
D. \{opt\_d\}\newline

Return exactly one uppercase letter only: A, B, C, or D.
\\
\midrule

image\_note (VS)
&
One image is provided: vehicle-side (ego) view.
\\
\midrule

image\_note (IS)
&
One image is provided: infrastructure-side (RSU) view.
\\
\midrule

image\_note (CO)
&
Two images are provided in order: (1) vehicle-side (ego) view, (2) infrastructure-side (RSU) view.
\\

\bottomrule
\end{tabular}
\end{table}

\normalsize

\printcredits


\bibliographystyle{cas-model2-names}

\bibliography{cas-refs}

\end{document}